\documentclass[]{opendrivelab}

\usepackage[toc,page,header]{appendix}
\usepackage[numbers]{natbib}
\usepackage{multicol}
\usepackage{graphicx}
\usepackage{float}
\usepackage{amsmath}
\usepackage{amssymb}
\usepackage{array}
\usepackage{enumitem}
\usepackage{booktabs}
\usepackage{makecell}
\usepackage{multicol,multirow}
\usepackage{setspace}
\usepackage{makecell}
\usepackage{fontawesome}
\usepackage[dvipsnames]{xcolor}
\usepackage{tasks}
\usepackage{pifont}
\usepackage{colortbl}
\usepackage{xcolor}
\usepackage{framed}
\usepackage{wrapfig}  
\usepackage{titletoc}
\usepackage{color} 
\usepackage{caption, subcaption, overpic, textpos}
\usepackage{titlesec}
\usepackage{comment}

\makeatletter
\DeclareRobustCommand\onedot{\futurelet\@let@token\@onedot}
\def\@onedot{\ifx\@let@token.\else.\null\fi\xspace}

\makeatother

\usepackage{xspace}
\newcommand{\modelname}{Compass 1\xspace}

\definecolor{deemph}{gray}{0.6}

\definecolor{baselinecolor}{gray}{.9}
\definecolor{yellow}{RGB}{218,165,32}
\definecolor{lightcyan}{rgb}{0.88, 1.0, 1.0}
\definecolor{lightskyblue}{rgb}{0.53, 0.81, 0.98}
\definecolor{aliceblue}{rgb}{0.94, 0.97, 1.0}
\definecolor{LightSlateBlue}{RGB}{70,130,180}
\definecolor{DeepBlue}{RGB}{65,100,170}
\definecolor{DeepPurple}{RGB}{136,105,160}
\definecolor{LightGreen}{RGB}{59,125,35}
\definecolor{LightRed}{RGB}{234,66,53}
\definecolor{cvprblue}{rgb}{0.21,0.49,0.74}

\usepackage{minitoc}

\title{%
  \raisebox{-0.2\height}{%
    \includegraphics[height=2.0em, trim=0px 0px 0px 0px, clip]{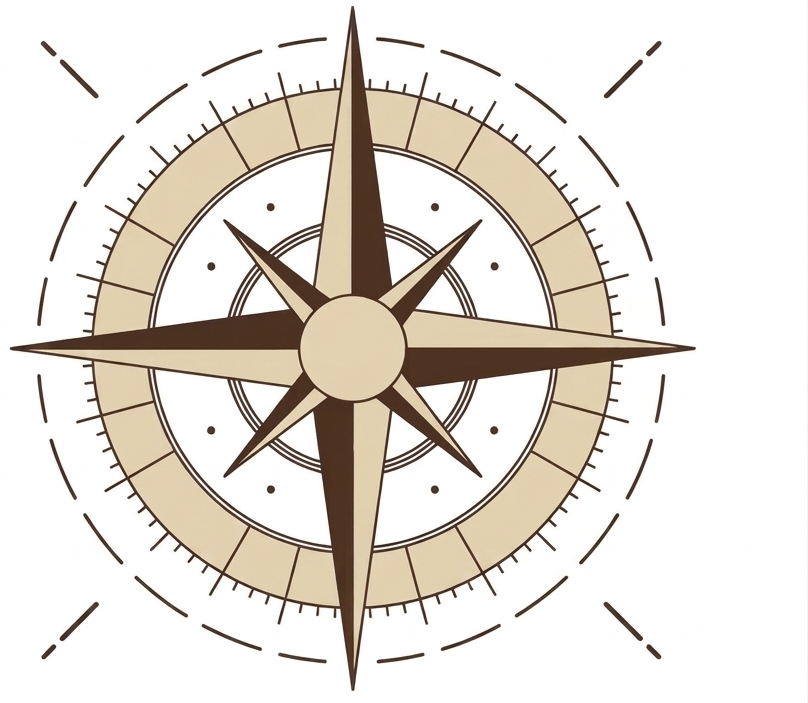}%
  }%
  \hspace{0.1em}Compass 1: Towards Open-World and Persistent Embodied Navigation from Monocular RGB Videos%
}

\author{
 Hai Zhang$^{1,*}$,
 Siqi Liang$^{1,*}$,
 Yanjun Guo$^{1}$, 
 Shoufeng Zhang$^{1}$ 
 Yunhao Wang$^{1}$, \\
 Hangyu Du$^{1}$, 
 Haoguang Mai$^{1}$, 
 Hongchen Li$^{2}$,
 Jiazhi Yang$^{3}$,
 Jialong Zeng$^{3}$ \\
 Yu Hu$^{4}$, 
 Chen Chen$^{5}$,
 Gangcheng Jiang$^{3}$, 
 Zishun Zhuo$^{1}$,
 Lirui Zhao$^{3}$\\
 Zhigang Li$^{5}$, 
 Yu Qiao$^{2}$, 
 Zhenjie Yang$^{1,3,\dagger}$, 
 Fu Zhang$^{1,4,\dagger}$, 
 Hongyang Li$^{3}$
}

\affiliation[1]{The University of Hong Kong}
\affiliation[2]{Shanghai Innovation Institute}
\affiliation[3]{Archon Robotics}
\affiliation[4]{Spatial Robotics} 
\affiliation[5]{Huawei Inc.}

\abstract{
Embodied navigation requires agents to achieve persistent spatial awareness and long-horizon execution stability in open-world environments. However, existing navigation architectures struggle to sustain both capabilities, remaining effective only over a handful of recent frames. 
We present \textbf{\modelname}, the first embodied navigation foundation model to unify language understanding and video generation while enabling memory backtracking through nearly \textbf{\textit{two minutes}} of visual history.
Without altering the underlying architecture, we introduce 
History Representation, which integrates sparse visual 
keyframes with an explicit History Representation Map to 
preserve long-term spatiotemporal context. To effectively 
leverage this history, we further introduce 
History Utilization, a structured VQA supervision
protocol comprising three categories: task-stage grounding, exploration consistency, and spatiotemporal target retrieval, enabling map-grounded reasoning and stable persistent long-horizon navigation.
To fuel this system, we establish a complete annotation pipeline for monocular RGB videos and open-source a massive dataset featuring over 500 hours of high-quality real-world interactions and over 2,000 hours of annotated web videos.
Trained mainly on the distribution of 1.0 meter (1m) height camera, \modelname generalizes seamlessly across diverse embodiments and scenarios, ranging from indoor environments to pedestrian-dense streets, spanning wheeled robot at 0.35m, quadruped dog at 0.5m, humanoid at 0.85m, and aerial drone at 1.5m.
\modelname achieves a dominant overall success rate of $79.4\%$ across open-world evaluations, compared to $17.8\%$ for the strongest baseline. 
It demonstrates exceptional zero-shot execution exceeding $90\%$ on foundational tasks, a $250\%$ boost in exploration, and an unparalleled leap from $0\%$ to $58\%$ in challenging persistent history backtracking.

}

\contribution[\star]{Equal contribution}
\contribution[\dagger]{Equal advising}
\date{\today}
\correspondence{Zhenjie Yang at \email{yangzj@hku.hk}, Fu Zhang at \email{fuzhang@hku.hk}}
\checkdata[Project Page]{\url{https://www.archon.tech/research/compass1}}

\begin{document}
\maketitle

\newpage

\begin{figure}[t]
    \centering
    \includegraphics[width=\linewidth]{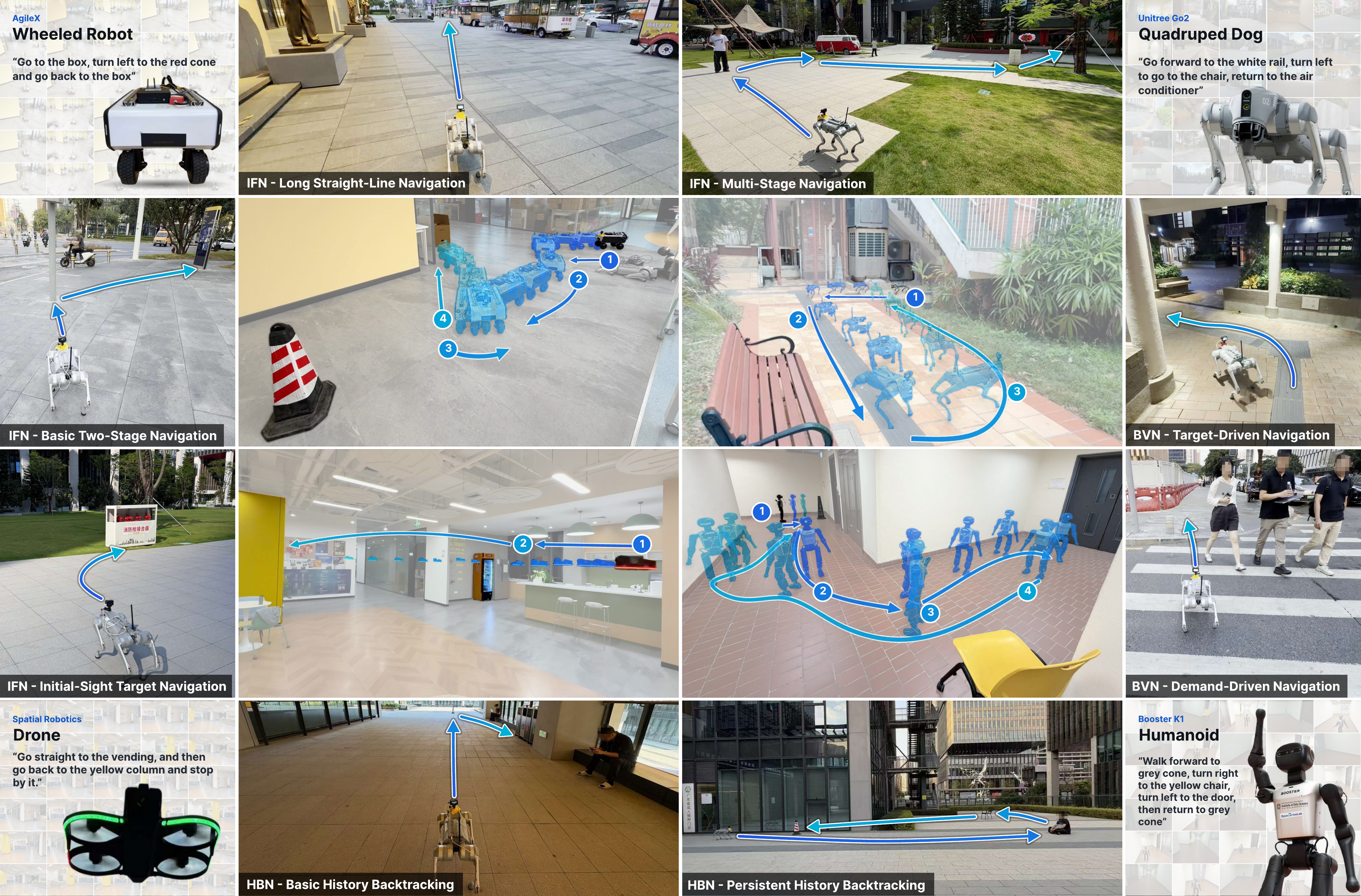}
    \vspace{-15pt}
    \caption{\textbf{Compass 1 Open-world and persistent embodied navigation.} We systematically design open-world evaluation settings tailored to core navigation tasks, comprising instruction-following navigation (IFN), beyond-the-view navigation (BVN), and history backtracking navigation (HBN). The outer panels illustrate reference navigation paths prescribed by task specifications.
    The inner panels depict sequential trajectory and pose overlays of zero-shot real-world executions achieved by \modelname across diverse robotic embodiments, including wheeled robot, quadruped dog, humanoid, and drone with camera heights of $0.35$\,m, $0.5$\,m, $0.85$\,m, and $1.5$\,m, respectively.}
    \vspace{-8pt}
    \label{fig:teaser}
\end{figure}

\section{Introduction}

Embodied navigation in unstructured, open-world environments requires physical agents to dynamically perceive visual signals, reason over high-level task instructions, and generate precise motor controls. 
Historically, the field has been dominated by the traditional training-free paradigm~\citep{wu2024voronav, rgpt2024, instructnav2025}. 
These systems rely heavily on specialized hardware, such as LiDAR sensors~\citep{sayplan2023, l3mvn2023} for dense point-cloud acquisition and depth cameras~\citep{vlfm2024, gu2024conceptgraphs} for spatial geometry, to reconstruct explicit topological maps onto which off-the-shelf Vision-Language Models (VLMs) are plugged in for decision-making.
Despite their modularity, these systems suffer from severe error accumulation and spatial degradation when confronting complex dynamic scenes, exposing a fragile generalization ceiling. 
More recently, emerging agentic navigation paradigms have attempted to execute decision-making via harness~\citep{zhang2026qwen} or pre-defined skill primitives~\citep{zhang2026supernav}. 
Yet, their practical deployment remains significantly bottlenecked by high inference latency, leaving them at a preliminary stage for physical deployment. 
In stark contrast, large-scale training-based navigation paradigm~\citep{zhang2024uni, zhang2026sparse} offers a far more effective alternative -- by unifying perception, reasoning, and execution within a monolithic and streamlined learning formulation.

Notwithstanding the impressive metrics achieved by recent large-scale training-based navigation approaches~\citep{awarevln2026, navida2026, wang2026lightnav, huang2026wnm} on simulated benchmarks~\citep{vlnce2020}, a pronounced disconnect emerges in physical deployment: prior arts overfit to simulation excessively and fail to generalize to real-world environments.
For instruction-following navigation tasks (IFN)~\citep{vlnce2020}, these methods frequently exhibit erratic behaviors, such as premature stopping, reversed turning, drifting off the intended direction specified by language instructions, and failing to stop accurately even when the target is directly in sight. 
For beyond-the-view navigation tasks (BVN)~\citep{zhang2026sparse}, these models are prone to left-right turning deadlocks driven by intention inconsistency between consecutive inference steps. 
Worse still, existing long-horizon formulations typically either naively feed the entire uncompressed visual history into the backbone~\citep{huang2026wnm} or apply heuristic, non-uniform compression~\citep{wang2026lightnav}. 
We observe that both strategies yield merely spurious context usage, failing to preserve granular spatiotemporal topology and rendering true history backtracking navigation (HBN) completely unachievable. 
These pervasive failure modes underscore the fundamental challenges in building a truly persistent, robust embodied navigation system in the real-world.

To resolve these deficiencies, we introduce \textbf{\modelname}, a navigation foundation model that addresses open-world and persistent embodied navigation without architectural modifications. 
For long-horizon history representation, we replace full visual streams with sparsely sampled visual keyframes as input and render their associated actions into a History Representation Map (HRM), which is processed natively by the backbone as an image input. 
We further design three visual-question-answering (VQA) categories encompassing task-stage grounding, exploration consistency, and spatiotemporal target retrieval, thereby fully unleashing the history utilization capabilities. 
To better incorporate spatial information and alleviate the short-horizon supervision problem inherent to language backbones~\citep{zhang2026sparse}, we naturally house these mechanisms within a unified understanding and generation backbone.

Beyond algorithmic innovations, a primary bottleneck to achieving robust open-world navigation stems from the scarcity of high-quality training data.
While scaling monocular RGB videos offers a promising avenue to bridge this gap~\citep{zhang2026sparse}, raw video clips inherently lack metric action trajectories, language instructions, and the specialized VQA annotations, making the whole labeling process prohibitively expensive. 
To overcome this bottleneck, we establish a semi-automated annotation pipeline designed for minimal human overhead. 
Human annotators are tasked solely with verifying language accuracy at stage boundaries and inspecting visualized action trajectories to filter out severe noise, significantly reducing human labor and annotation cost.
Leveraging this pipeline, we curate a massive navigation corpus comprising over 500 hours of high-quality real-world data and 2,000 hours of annotated web-scale video data, providing an unprecedented multi-modal foundation to fuel open-world embodied navigation.

We structure our real-world evaluation benchmark around essential navigation capabilities across highly diverse physical scenarios. 
Across these evaluations, \modelname achieves exceptional zero-shot execution exceeding $90\%$ on foundational tasks, including initial-sight target navigation, basic two-stage navigation, and basic history backtracking. 
In exploration tasks covering target-driven navigation and demand-driven navigation, \modelname achieves an average success rate of $84\%$, delivering a $250\%$ performance improvement over the strongest baseline. 
Most notably, across challenging long straight-line navigation, multi-stage navigation, and persistent history backtracking, \modelname demonstrates an unparalleled performance leap, boosting the average success rate from $5\%$ to $60\%$.
Crucially, despite being trained primarily on a single camera height distribution at 1.0 meter (1m), \modelname exhibits exceptional zero-shot transferability across varied embodiments, smoothly adapting to distinct camera heights, ranging from wheeled robots at 0.35m and quadrupeds at 0.5m to humanoids at 0.85m and aerial drones at 1.5m.

\subsection{Research Questions}

The pivotal gist in this work boils down to -- \textit{How could a model excel at 
executing open-world and persistent navigation while generalizing across diverse scenarios and embodiments?}

This breaks into three concrete questions:
\begin{itemize}
    \item How can we efficiently represent navigation history such that the policy can pinpoint previously observed targets, track current task stages, and discern how to navigate back over long horizons?
    \item How can we utilize this history representation to support diverse navigation tasks and learn a generalizable policy from limited high-quality data?
    \item How can we establish an effective annotation pipeline for raw video corpora to significantly reduce human labor and resource overhead?
\end{itemize}

\subsection{Contributions}

The main contributions are:
\begin{itemize}
    \item \textbf{Keyframe-Level Unified Backbone \& History Representation Map:} To our knowledge, we present the first keyframe-level foundation model built on a unified backbone for embodied navigation. By condensing keyframes and actions into a compact HRM, our method grants the agent persistent spatiotemporal reasoning and nearly two-minute memory backtracking capabilities with zero architectural modification.
    \item \textbf{Structured VQA Supervision Protocol:} We design three structured VQA categories to fully utilize the history reasoning capability and effectively resolve physical execution failure modes including premature stopping, reversed turning, direction drifting, left-right turning deadlocks, and spurious history utilization.
    \item \textbf{Cost-Effective Data Annotation Pipeline:} We establish a semi-automated annotation pipeline requiring human oversight solely at stage boundaries and trajectory noise filtering, enabling the release of a massive navigation corpus featuring over 500 hours of real-world interactions and over 2,000 hours of annotated web videos.
    \item \textbf{Thorough Physical Evaluation \& Zero-Shot Cross-Embodiment Generalization:} Extensive real-world experiments demonstrate that \modelname drastically outperforms strong baselines across all physical navigation tasks. Remarkably, despite being trained mainly on a fixed camera height distribution, it generalizes zero-shot across diverse embodiments, and substantial camera height variations.
\end{itemize}

\section{Methodology}

\begin{figure}[ht]
    \centering
    \includegraphics[width=\linewidth, trim=0px 0px 0px 0px, clip]{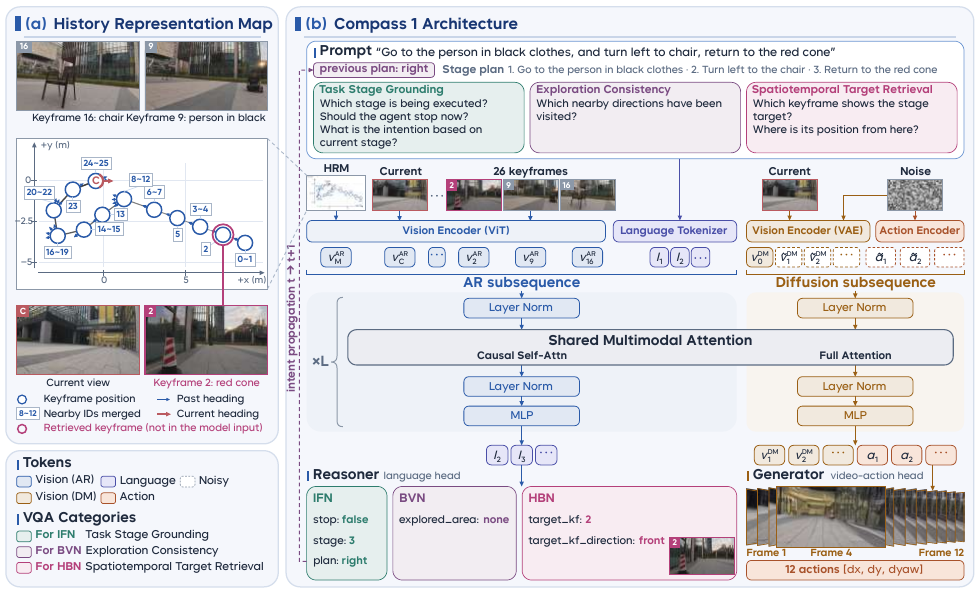}
    \vspace{-15pt}
    \caption{\textbf{Overview of the history representation map and the architecture of \modelname.} 
    (a) The history representation map (HRM) projects past keyframes and history trajectories onto a 2D visual image with dynamically aggregated indices and heading arrows to encode dense spatial topology. 
    (b) The architecture of \modelname leverages a unified multimodal understanding and generation backbone to process prompt instructions, the HRM, and keyframe sequences. 
    Structured VQA supervision categories enforce history-aware reasoning while a diffusion-based video-action head continuously generates low-latency action trajectories.}
    \vspace{-4pt}
    \label{fig:method}
\end{figure}

The design of \modelname is guided by two fundamental pillars: history representation and history utilization (\Cref{sec:history_repr,sec:history_util}), alongside the synergistic orchestration of multimodal understanding and generation (\Cref{sec:sys_design}).

The whole architecture and HRM are shown in~\Cref{fig:method}.

\subsection{History Representation}
\label{sec:history_repr}
Given the intrinsic scarcity of high-quality, real-world navigation data, a primary design philosophy of \modelname is to preserve the pre-trained foundation backbone without introducing any architectural modifications. 
To achieve this, our history representation is built upon two core aspects: keyframe-level sparsification and history representation map construction.

\textbf{Keyframe-Level Sparsification:} 
Existing methods face inherent trade-offs in long-horizon history modeling. 
Naively appending all visual inputs severely degrades attention quality with heavy computational overhead~\citep{zhang2024navid}. 
Uniformly sampling visual frames at a fixed frequency to fit a limited context window leads to severe information loss~\citep{cheng2025navila}, whereas employing specialized pooling mechanisms inevitably introduces extra network modules~\citep{zhang2024uni, wang2026lightnav}.
To circumvent these limitations, we summarize visual context through action-guided keyframe selection. 
This strategy effectively isolates critical spatial observations without adding structural overhead.
Details of keyframe selection are deferred to \Cref{sec:keyframe_select}.

Crucially, deploying a policy trained on keyframe sequences proves essential during inference for mitigating the impact of control uncertainty across diverse embodiments. 
Particularly for embodiment like humanoids, micro-actions often fail to overcome physical execution limits to produce actual physical movement, rendering continuous visual inputs non-uniform and inducing severe out-of-distribution (OOD) shifts relative to continuous training video frames, which inevitably degrades policy performance. 
We find that simply triggering keyframe selection at inference time based on fixed distance and angle thresholds effectively addresses this issue and yields impressive navigation performance.

\textbf{History Representation Map Construction:}
For tasks requiring true history utilization, such as HBN, the policy must not only pinpoint previously observed targets but also discern how to physically navigate back to them. 
Over such extended horizons, language-based summaries fail to adequately capture subtle scene information and historical action trajectories~\citep{torne2026memmultiscaleembodiedmemory}. 
Inspired by explicit map construction in training-free paradigms~\citep{wu2024voronav, instructnav2025} and trajectory visualization during data annotation, we format past action trajectories into a 2D visual canvas, termed the \textit{History Representation Map (HRM)}. 
Specifically, keyframes are projected onto the HRM canvas and labeled with their corresponding frame indices. 
To resolve spatial clutter in densely visited regions, spatially proximal keyframe indices are dynamically aggregated (e.g., denoted as ``2--5'' to indicate multiple keyframes within the localized neighborhood). 
Furthermore, since video streams possess an intrinsic temporal directionality, directional arrows are rendered along the trajectory to encode past headings. 
The current step is explicitly highlighted with a red letter \textcolor{red}{C} accompanied by a red arrow specifying its heading.
By directly feeding the generated HRM alongside keyframes into the backbone as native image inputs, \modelname achieves long-horizon spatiotemporal reasoning with zero architectural modification. 
Under this design, while the visual sequence fed into the backbone remains compact and keyframe-sparse, the HRM provides dense, explicit topological grounding over the entire navigation trajectory.

\subsection{History Utilization}
\label{sec:history_util}

To maximize the efficacy of historical contexts without architectural modifications, the optimal strategy is to intrinsically embed spatiotemporal reasoning capabilities into the foundation backbone. 
We achieve this by formulating a structured VQA supervision protocol. 
Empirically, we observe that naively taking the HRM as an input causes the backbone to ignore this guidance. 
Thus, our VQA protocol serves as an essential \textit{map activation mechanism}. 
To resolve pervasive execution failures across Instruction-Following Navigation (IFN), Beyond-the-View Navigation (BVN), and History Backtracking Navigation (HBN), we design three specialized categories of VQA objectives:

\textbf{Task-Stage Grounding (for IFN):} 
Standard IFN policies frequently suffer from execution flaws, including premature stopping, direction mismatch, and failing to stop accurately even when the target is directly in sight~\citep{huang2026wnm, navida2026}. 
To rectify these issues, we first employ Qwen-3.8-27B~\citep{qwen38} to parse raw language instructions into structured task stages. 
The backbone is then queried to identify the current stage from the overall instruction breakdown, predict the immediate next-step high-level intention aligned with the current stage, and determine whether a termination criterion is met. 
Notably, to accelerate inference latency and allow sufficient execution buffer, our termination supervision is constructed anticipatorily $4$ steps prior to actual stopping (see \Cref{sec:sys_design}).

\textbf{Exploration Consistency (for BVN):} 
In open-world BVN, exploration trajectories inherently exhibit multimodal distributions where turning left, right, or moving straight are all plausible when the target is absent~\citep{zhang2026sparse, huang2026wnm}. 
As conventional approaches lack visibility into prior intentions, step-by-step inference becomes decoupled, leading to left-right turning deadlocks~\citep{huang2026wnm}. 
We mitigate this by querying the backbone on whether the left or right spatial regions have been previously visited, anchored on the HRM. 
Furthermore, we establish an intention propagation mechanism, passing the predicted high-level intention from the previous inference step into the successive prompt text, enforcing execution consistency throughout exploration.

\textbf{Spatiotemporal Target Retrieval (for HBN):} 
For long-horizon HBN, policies typically fail to extract actionable context from visual history. 
Combining the parsed task stage with the HRM, we design a joint spatiotemporal retrieval question. 
Given a target query, the backbone simultaneously identifies the historical keyframe index capturing the target and predicts the relative direction required to navigate toward that keyframe from the current position.
This general formulation not only resolves complex history backtracking but also naturally enhances target-recovery maneuvers when a previously observed target is inadvertently bypassed during ongoing navigation.

\subsection{System Design}
\label{sec:sys_design}
To seamlessly unify VQA reasoning with fine-grained spatiotemporal prediction, \modelname adopts Cosmos3-Edge-4B~\citep{agarwal2026cosmos} as its foundation backbone. 
To our knowledge, \modelname is the first framework in embodied navigation to leverage a unified multimodal understanding-and-generation architecture as its core backbone. 
Formally, the system takes the current keyframe-level observation stream, the HRM, and prompt instructions as native inputs, operating through two synergistic decoding heads: an autoregressive language head that handles structured VQA reasoning, alongside a diffusion-based action-video head that jointly generates the future and predicts action trajectories. 
\begin{figure}[ht]
    \centering
    \includegraphics[width=0.8\linewidth]{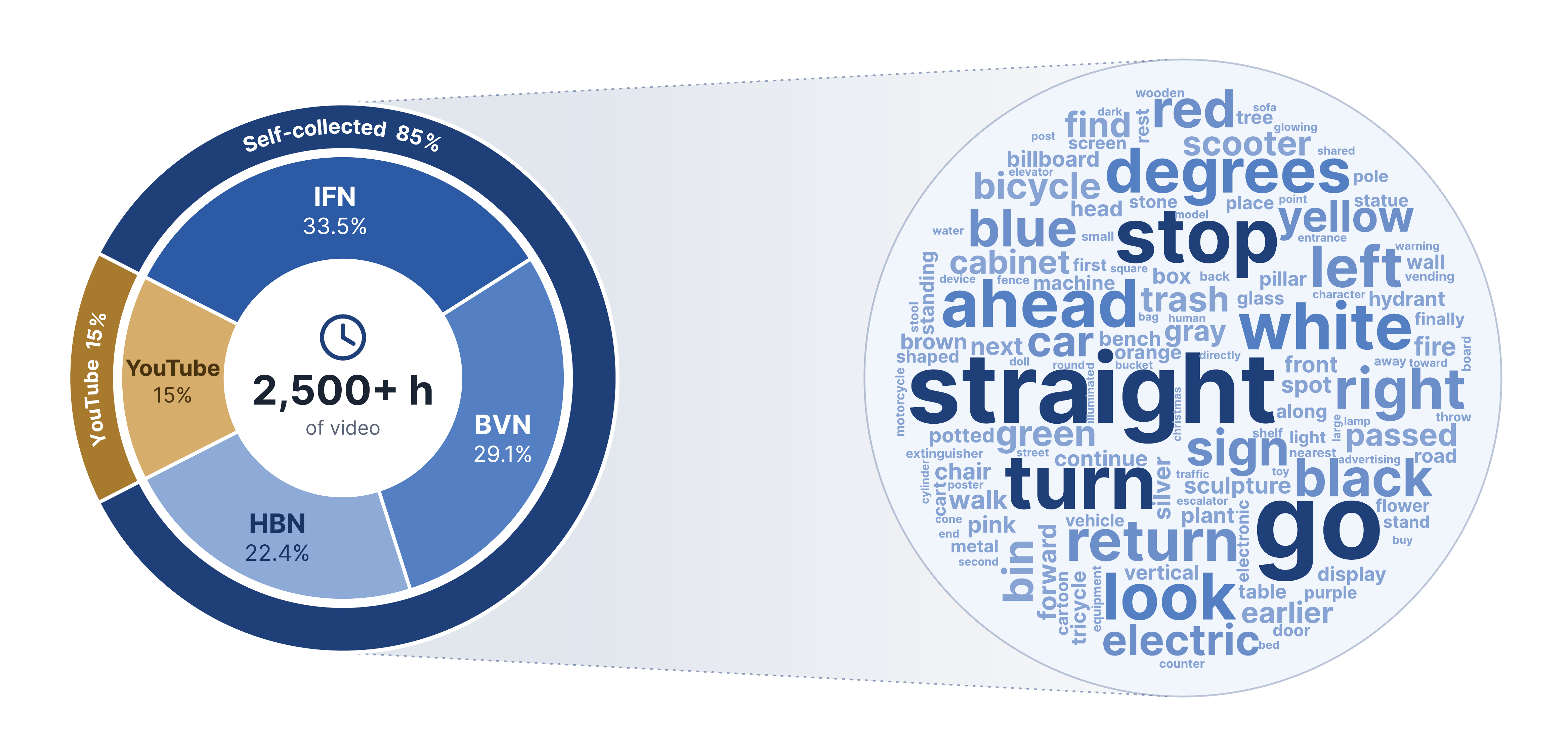}
    \vspace{-15pt}
    \caption{\textbf{Training data recipe of \modelname.} 
    The training corpus comprises over $2{,}500$ hours of video, integrating over $500$ hours self-collected real-world trajectories across IFN ($33.5\%$), BVN ($29.1\%$), and HBN ($22.4\%$), alongside over $2{,}000$ hours annotated web-scale YouTube videos. During training, we sample the self-collected data and YouTube data at a ratio of $85\%$ and $15\%$, respectively.
    The right panel depicts the word cloud visualization of synthesized navigation instructions, highlighting rich motion primitives and landmark semantics.}
    \vspace{-8pt}
    \label{fig:training_data_recipe}
\end{figure}

As joint training enforces strong VQA reasoning within the backbone, action generation at inference time can operate independently of the language head. 
Exploiting this architecturally decoupled nature, we adopt an asynchronous deployment strategy on a single GPU. 
Upon generating low-latency action trajectories, the system immediately streams them to the physical platform while asynchronously completing the language reasoning forward pass to extract intentions and the stopping signal.
Once a stopping signal is triggered, an asynchronous interrupt flag is transmitted to the onboard controller to immediately execute termination.
We further incorporate KV-cache for keyframe sequences and cuda graph compilation for diffusion denoising to accelerate model inference.

\subsection{Training Recipe}
\label{sec:training_recipe}

As illustrated in~\Cref{fig:training_data_recipe}, we co-train \modelname using a diverse dataset comprising over 500 hours of high-quality self-collected real-world data and 2,000 hours of annotated YouTube videos. 
To ensure that the model develops balanced and robust accuracy across diverse VQA queries without overfitting to specific prompts, we implement a selective query sampling strategy during training, where termination is queried continuously, while other queries are sampled with a 50\% probability per training sample.
The entire model is optimized end-to-end on a cluster of 32 NVIDIA H200 GPUs for a total of 60000 steps, using a global batch size of 256. 
To balance multimodal reasoning, video generation, and action prediction, we apply differentiated learning rate schedules across network heads: the visual head and language head maintain a constant learning rate of $1\times10^{-5}$, whereas the action head is initialized with a learning rate of $1\times10^{-4}$, warmed up over the first 1\% of total training steps, and subsequently decayed to $1\times10^{-5}$ via a cosine annealing schedule. 
The overall optimization objective is a joint loss weighted across task dimensions, with a loss ratio of $3:1:1$ for the VQA loss, video generation loss, and action prediction loss, respectively.

\section{Dataset Curation}
\label{sec:data_curation}

\begin{figure}[!ht]
    \centering
    \includegraphics[width=\linewidth, trim=0px 0px 0px 10px, clip]{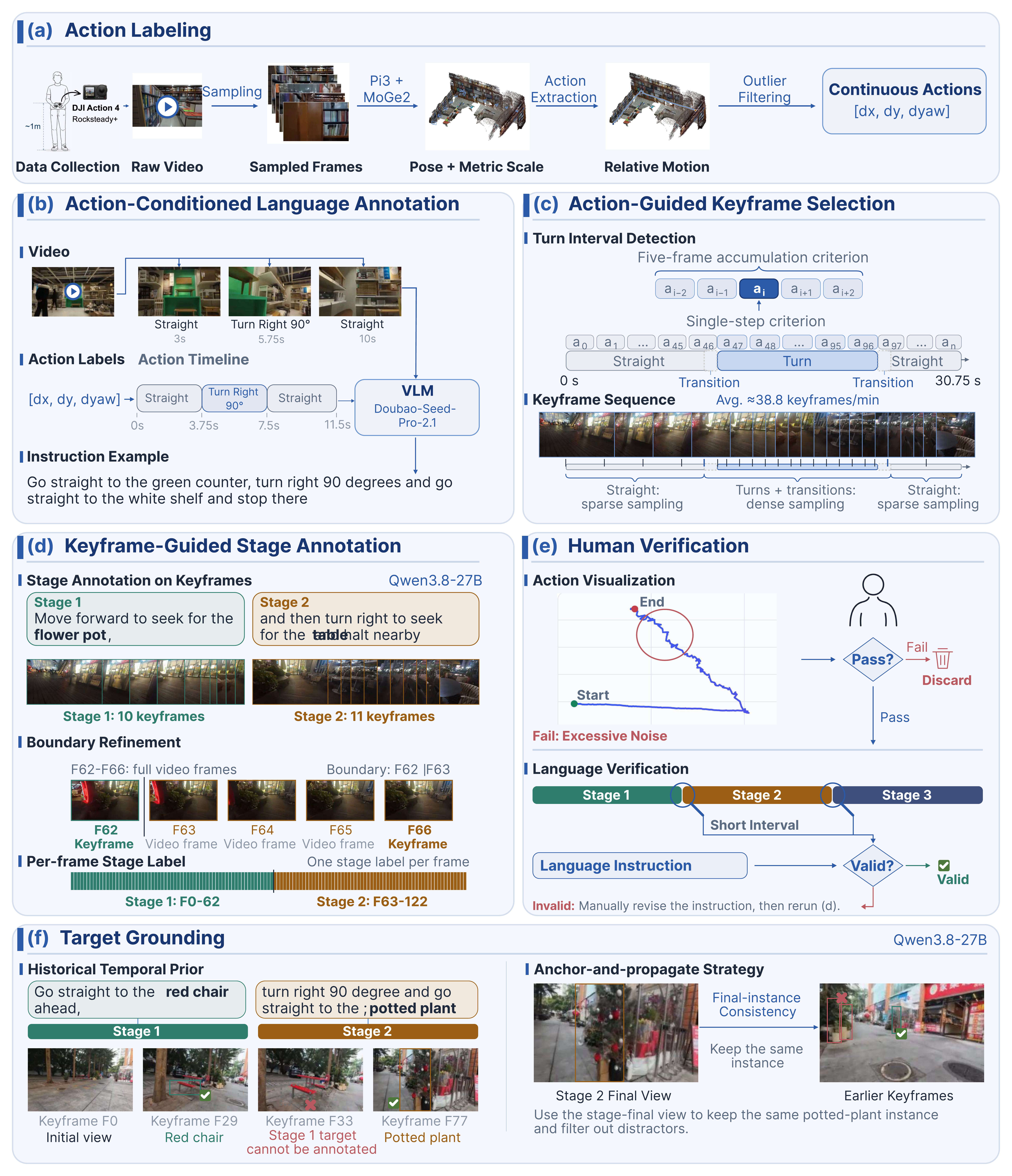}
    \vspace{-22pt}
    \caption{\textbf{Semi-automated dataset curation pipeline.} 
    The pipeline processes raw egocentric videos through six core stages. 
    (a) Action labeling extracts metric continuous trajectories while filtering out noisy outliers. 
    (b) Action-conditioned language annotation leverages action primitives as spatiotemporal priors to guide VLM in synthesizing structured instructions. 
    (c) Action-guided keyframe selection adaptively downsamples video streams through turning interval detection and non-uniform sampling. 
    (d) Keyframe-guided stage annotation decomposes tasks into target-oriented stages and refines temporal boundaries. 
    (e) Human verification streamlines quality control by inspecting trajectory action visualizations and short boundary intervals. 
    (f) Target grounding utilizes historical temporal priors and an anchor-and-propagate strategy to ensure consistent bounding box annotations across keyframe sequences.}
    \vspace{-10pt}
    \label{fig:data_curation}
\end{figure}

Owing to the notorious physical domain gaps and severe visual artifacts present in existing simulators~\citep{2019habitat}, relying on simulator-generated navigation data frequently degrades zero-shot real-world deployment. 
To circumvent data contamination from synthetic trajectories, \modelname adopts a real-world data collection philosophy inspired by \citet{zhang2026sparse}, scaling up physical navigation trajectories while aggregating web-scale video data from YouTube.

Specifically, we collect high-fidelity egocentric video streams using DJI Action 4 cameras with hardware stabilization enabled. 
Operators adhere to strict protocol guidelines to systematically record trajectories across three core navigation tasks: IFN, BVN, and HBN. 
BVN is subdivided into target-driven exploration and demand-driven exploration.

Both self-collected real-world trajectories and YouTube videos are processed through our semi-automated annotation pipeline to extract action labels, language instructions, keyframes, and VQA supervisions, as illustrated in \Cref{fig:data_curation}. 
Given the significantly larger volume of YouTube data, human verification is exclusively applied to our self-collected trajectories.

\subsection{Action Labeling from Monocular RGB Videos}
\label{sec:action_labeling}

Following the paradigm of NavDreamer~\citep{navdreamer2026}, we extract metric continuous action trajectories from monocular RGB videos by combining $\pi^3$~\citep{wang2026pi} and MoGe-2~\citep{2025moge2}, as shown in \Cref{fig:data_curation}(a). 
However, trajectory visualizations reveal that raw extracted actions frequently contain outliers induced by monocular ambiguity and visual noise. 
These anomalies manifest in three primary forms:
\begin{itemize}
    \item \textbf{Abnormal action scales,} where predicted action scale far exceed the nominal per-step physical movement during real-world collection.
    \item \textbf{Excessive roll and pitch variations,} which typically signify camera stabilization failures or severe jitter from human operators.
    \item \textbf{Erratic multi-axis spikes,} characterized by simultaneous large variations in $\text{dyaw}$, $\text{dx}$, and $\text{dy}$ during tight-radius cornering maneuvers near obstacles.
\end{itemize}
To ensure the integrity of our imitation learning corpus, trajectories exhibiting any of these failure modes are filtered out from the training dataset.

\subsection{Action-Conditioned Language Annotation}
\label{sec:language_labeling}

Directly prompting VLM with raw video streams to synthesize navigation instructions frequently suffers from hallucinated temporal boundaries, direction misidentification, and stage omission. 
To mitigate these limitations, we introduce an action-conditioned language annotation protocol that leverages the extracted metric trajectories as explicit spatiotemporal priors.

Continuous action sequences are first aggregated along the temporal timeline to partition the whole video into discrete, semantically meaningful action primitives, including \textit{moving straight}, \textit{turning left}, \textit{turning right}, and \textit{searching}. 
In the pipeline detailed in~\Cref{fig:data_curation}(b), this structured action timeline, alongside the raw video stream, is fed into Doubao-Seed-Pro-2.1~\citep{team2026seed2}. 
Guided by these deterministic motion primitives, the VLM reliably grounds visual landmarks (e.g., trash can or billboard) to their corresponding trajectory segments, ultimately synthesizing fine-grained, stage-aligned instructions with guaranteed structural consistency.

These synthesized instructions serve as the foundational annotations, customized to align with specified task categories. 
IFN and target-driven BVN directly deploy the raw generated instructions. 
For demand-driven BVN, the target description in the searching stage is reformulated into high-level user demands. 
For HBN, which mandates a backtracking at the last stage during data collection, only the final clause is replaced to direct the agent back to its target.

\subsection{Action-Guided Keyframe Selection}
\label{sec:keyframe_select}

Processing dense video frames indiscriminately introduces redundant spatial contexts and excessive computational overhead. 
To address this, we propose an action-guided keyframe selection strategy that adaptively downsamples video streams using trajectories as spatio-temporal priors.

As illustrated in~\Cref{fig:data_curation}(c), candidate turning intervals are first identified by detecting localized heading variations along the trajectory. 
Turning evidence is captured through a dual-condition criterion based on step-wise yaw changes, which evaluates both an instantaneous single-step threshold and a five-frame yaw accumulation window anchored at a central frame. 
This formulation effectively filters out minor heading noise and oscillatory motion. 
Nearby turning steps are subsequently merged into contiguous motion segments with bidirectional temporal buffers to preserve critical pre- and post-turn visual transitions.

Based on these extracted motion semantics, keyframes are sampled under a non-uniform priority scheme where the initial frame is strictly retained as the first keyframe. 
Dense sampling is applied across turning and transition regions whenever the yaw accumulation exceeds $30^\circ$ to prevent missing target objects during rapid rotation, whereas straight segments are sampled sparsely at $\text{3s}$ intervals to eliminate background redundancy.
Through this dynamic downsampling strategy, a dense one-minute video stream (240 frames at 4 FPS) is efficiently compressed to an average of only 40 keyframes, significantly reducing computational overhead while retaining essential navigation semantics. 

\subsection{Keyframe-Guided Stage Annotation}
\label{sec:stage_anno}

Directly feeding dense video streams into VLM is computationally prohibitive and prone to temporal misalignment. 
To overcome this limitation, stage annotation operates on the pre-extracted sparse keyframes rather than raw video frames. 
As illustrated in \Cref{fig:data_curation}(d), the complete navigation instruction and keyframe sequence are jointly fed into Qwen-3.8-27B~\citep{qwen38} to decompose the task into target-oriented stages and assign the stage to each keyframe. 

Since the initial keyframe mapping inevitably yields coarse temporal boundaries, a fine-grained refinement step is executed across boundary intervals.
We then use Qwen-3.8-27B to inspect the interval containing raw video frames between the boundary keyframes and their immediate successors, precisely pinpointing the exact frame where the subgoal is fulfilled.
Building upon these refined temporal boundaries, per-frame stage labels are efficiently propagated across the entire continuous video stream. 

\subsection{Human Verification}
\label{sec:human_verify}

To ensure high dataset quality while retaining scalability, we incorporate a lightweight human-in-the-loop verification protocol. 
Unlike conventional annotation pipelines that require annotators to watch full video streams, manually craft navigation instructions, and label frame-level targets~\citep{zhang2026sparse}, we streamline human intervention into two highly efficient inspection tasks.

As illustrated in~\Cref{fig:data_curation}(e), annotators directly inspect the action visualization to identify anomalous artifacts within metric trajectories. 
Trajectories exhibiting excessive motion noise are directly discarded to guarantee training stability. 
Subsequently, language alignment is verified using only short video intervals surrounding stage boundaries rather than full-length sequences. 
Annotators directly verify whether the target objects and physical movements within these boundary intervals remain consistent with the language instruction. 
If an inconsistency is identified, the instruction is manually corrected, after which the trajectory re-enters the stage annotation pipeline described in \Cref{sec:stage_anno} to update temporal stage assignments.

\subsection{Target Grounding}
\label{sec:target_ground}

To establish fine-grained spatiotemporal alignment, we further utilize Qwen-3.8-27B~\citep{qwen38} to ground semantic targets by annotating bounding boxes across all extracted keyframe sequences guided by the language instruction. 
Directly detecting targets frame-by-frame often introduces redundant candidates and temporal inconsistencies. 
To guarantee high-quality grounding, we apply two trajectory-level structural constraints during annotation.

As depicted in \Cref{fig:data_curation}(f), we enforce a historical temporal prior where target annotations exist exclusively within their corresponding stage and all preceding stages.
If a target object from an earlier stage reappears in subsequent stages, its bounding box annotations in the later stages are explicitly removed to prevent semantic ambiguity. 
Furthermore, to eliminate candidate redundancy caused by multiple visual instances, we adopt an anchor-and-propagate strategy. 
The canonical target bounding box is uniquely identified at the final keyframe of each stage and subsequently propagated backward through all preceding keyframes. 
This backward propagation strategy filters out noisy candidates and ensures robust visual grounding across the entire navigation trajectory.

\section{Experiments}
\label{sec:experiments}

To thoroughly evaluate \modelname, we conduct extensive experiments focusing on open-world embodied navigation capabilities. 
\Cref{sec:eval_protocol} outlines the evaluation protocol for zero-shot real-world deployment across diverse scenarios and embodiments. 
\Cref{sec:qualitative_results,sec:quantitative_results,sec:diverse_robots} demonstrate the cross-scenario and cross-embodiment capability of \modelname. 
Detailed ablation studies (\Cref{sec:ablation_studies}) and inference efficiency analysis (\Cref{sec:inference_efficiency}) then validate the rationality of our model design and its internal mechanisms. 
Finally, comprehensive simulation evaluation results comparing various SOTA baselines are reported on established benchmarks (\Cref{sec:simu_results}).

\subsection{Evaluation Protocol}
\label{sec:eval_protocol}

To rigorously demonstrate the edge of \modelname, we select four representative SOTA baselines spanning distinct architectural paradigms for comparative analysis.

\textbf{VLM-Based Paradigm:}
\begin{itemize}
    \item \textbf{NaVIDA}~\citep{navida2026} augments policy learning with action-grounded visual dynamics driven by inverse dynamics supervision.
    \item \textbf{AwareVLN}~\citep{awarevln2026} equips special tokens to dynamically alternate between reasoning and action modes.
    \item \textbf{LightNav-0}~\citep{wang2026lightnav} utilizes points as latent chain-of-thought representations.
\end{itemize}

\textbf{Video-Generation-Based Paradigm:}
\begin{itemize}
    \item \textbf{WNM-3D}~\citep{huang2026wnm} employs a mixture-of-transformer architecture to co-optimize video and action generation.
\end{itemize}

To ensure strict fairness across evaluations, all four baselines are fine-tuned on our dataset. 
We defer detailed implementations to~\Cref{sec:app_impl_real}.
To reflect the peak performance of each baseline, we report the higher success rate between its original open-sourced weight and its fine-tuned counterpart in our evaluation results.

We conduct zero-shot real-world evaluations across four representative indoor and outdoor scenarios, including \textbf{Indoors} (Office) and \textbf{Outdoors} (Terrace, Complex, and Street). 
These environments range from structured office rooms to highly dynamic street environments with moving pedestrians and vehicles, where the street scenario specifically evaluates the robustness against dynamic disturbances. 

To systematically assess diverse navigation skills, we construct a comprehensive evaluation suite comprising eight distinct tasks organized across three navigation categories.
To guarantee statistical rigor, each method undergoes 10 independent trials for every task across all four scenarios, accumulating \textbf{320} total evaluations per approach.
We report the \textbf{Success Rate} averaged by 10 trials for each task as the evaluation metric.
Detailed task setting is shown in~\Cref{sec:app_openworld_config}.

\begin{figure}[ht]
    \centering
    \includegraphics[width=\linewidth]{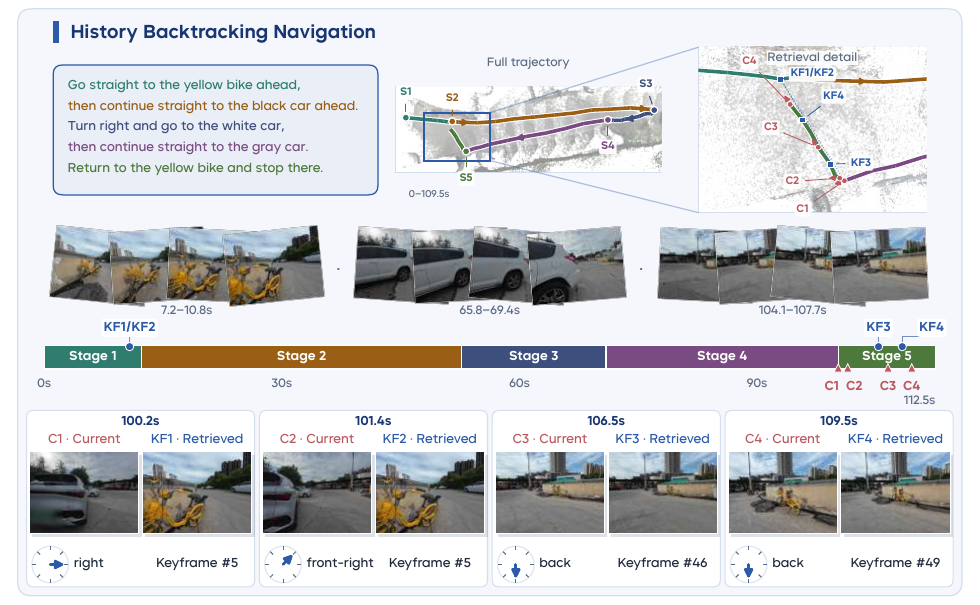}
    \caption{\textbf{Open-loop evaluation results on HBN.} Over a nearly two-minute trajectory, \modelname successfully identifies the keyframe index at approximately 8s and the direction corresponding to that keyframe at 100s, and continuously updates these estimates upon re-observing the target after 106s. Color coding across task stages (S1 to S5) connects the language instruction, temporal timeline, and estimated robot trajectory. The zoomed-in retrieval details and bottom image pairs show the current observations (C) alongside retrieved keyframes (KF), specifying the estimated target directions and keyframe indices.}
    \label{fig:open_loop}
\end{figure}

\begin{figure}[!ht]
    \centering
    \includegraphics[width=\linewidth]{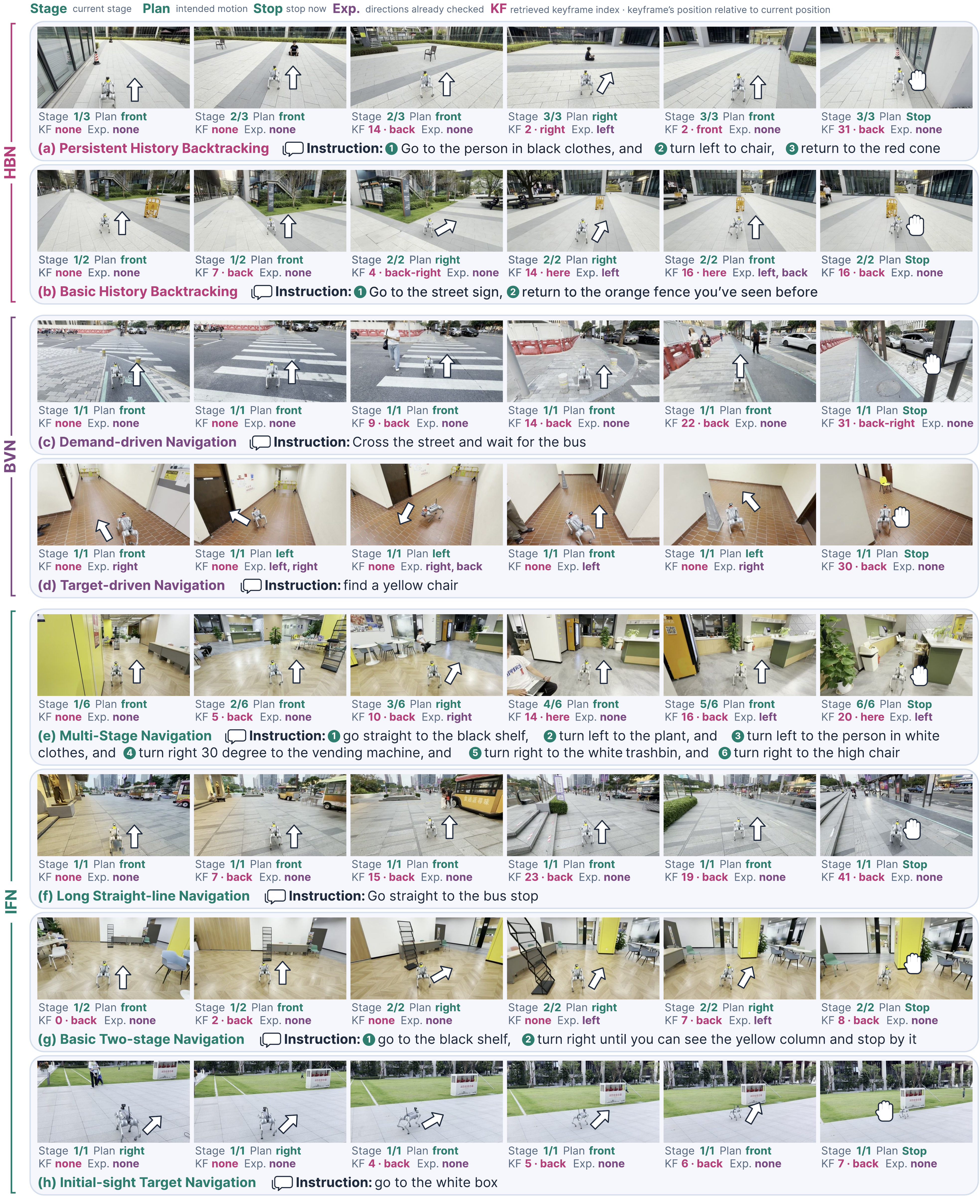}
    \vspace{-5pt}
    \caption{\textbf{Qualitative zero-shot results of \modelname in open-world environments across HBN, BVN, and IFN.}}
    \vspace{-5pt}
    \label{fig:closed_loop}
\end{figure}

\textbf{Instruction-Following Navigation Capabilities:}
\begin{itemize}
    \item \textbf{Initial-Sight Target Navigation}: The target object is initially visible within the field of view but not directly aligned with the heading direction. This task tests whether the agent can dynamically adjust its heading toward the target and stop accurately.
    \item \textbf{Basic Two-Stage Navigation}: The agent must navigate to an intermediate target before executing a turn, which may exceed $90^\circ$, to reach the final goal. This evaluates whether the policy can resume incomplete turns across multiple inference steps without making premature stops or directional errors.
    \item \textbf{Long Straight-Line Navigation}: The agent is required to maintain a consistent heading along a straight trajectory exceeding 30m, even when the target remains invisible for an extended period.
    \item \textbf{Multi-Stage Navigation}: The trajectory consists of at least four sequential stages. This task evaluates whether the agent can track multiple stages without confusing the turn order or stopping prematurely.
\end{itemize}

\textbf{Beyond-the-View Navigation Capabilities:}
\begin{itemize}
    \item \textbf{Target-Driven Navigation}: The agent navigates (semi-)enclosed areas, such as dead ends, to evaluate whether it avoids left-right turning deadlocks and successfully escapes from such restricted areas.
    \item \textbf{Demand-Driven Navigation}: The agent is required to combine target-driven navigation with semantic demand reasoning to determine appropriate stopping locations in (semi-)enclosed areas.
\end{itemize}

\textbf{History Backtracking Navigation Capabilities:}
\begin{itemize}
    \item \textbf{Basic History Backtracking}: The revisited target is located adjacent to the target of the preceding stage, becoming visible either immediately upon turning around or after a 180$^\circ$ turn followed by a short straight distance along a near-linear trajectory. Given that typical baselines strongly rely on immediate history frames, this tests whether they effectively exploit these recent spatiotemporal cues to re-localize the target.
    \item \textbf{Persistent History Backtracking}: The revisited target is entirely occluded from the target of the preceding stage. This tests the capability for long-horizon spatiotemporal reasoning and historical trajectory backtracking.
\end{itemize}

Physical validation primarily leverages the Unitree Go2 \textbf{Quadruped Dog} shown in~\Cref{fig:teaser}. 
The physical deployment of all evaluated methods relies on odometry to execute closed-loop velocity control.
We utilize a DJI Action4 camera mounted to the quadruped head via adhesive bonding at roughly 0.5m above ground to capture the egocentric video stream. 
The dynamic locomotion of the robot introduces continuous body jitter that induces pitch angle variations during deployment. 
Combined with the 0.5m mounting height that significantly deviates from the training distribution, this hardware configuration naturally establishes a challenging real-world testbed for generalization.
Cloud-side processing powered by NVIDIA RTX PRO 6000 GPU handles remote inference for both our method and the baselines.

We further assess cross-embodiment scalability of \modelname by extending real-world testing to three additional embodiments spanning different camera heights. 
These comprise an AgileX \textbf{Wheeled Robot} at 0.35m, a Booster K1 \textbf{Humanoid} at 0.85m, and a Spatial Robotics \textbf{Drone} at 1.5m. 
All ground platforms adopt the DJI Action4 camera, whereas the Drone operates with its native industrial camera.

\begin{figure}[!ht]
    \centering
    \includegraphics[width=\linewidth]{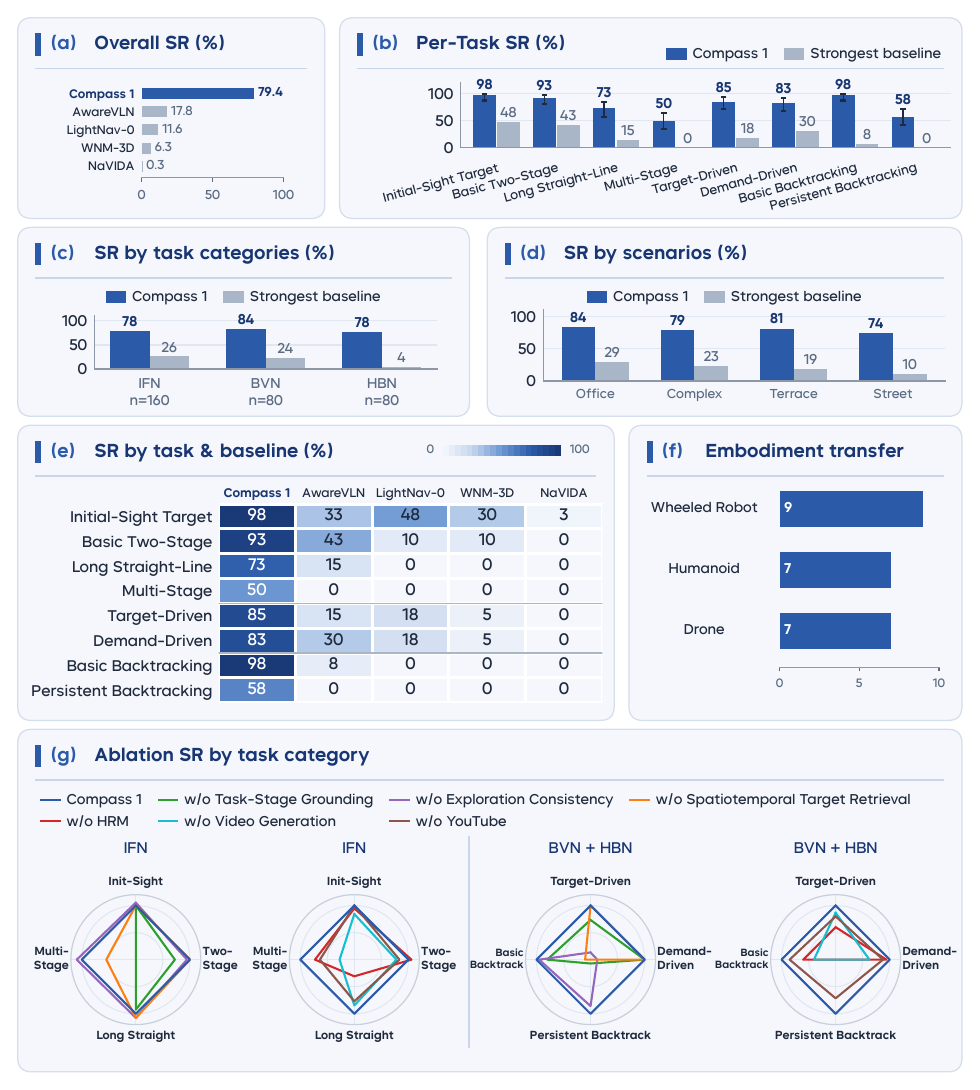}
    \caption{\textbf{Quantitative zero-shot results of \modelname against baselines and different variants in open-world environments across HBN, BVN, and IFN.} 
    (a) Overall success rate across all open-world tasks. 
    (b) Per-task success rate comparing \modelname with the strongest baseline. 
    (c) Success rate breakdown by task categories including IFN, BVN, and HBN. 
    (d) Success rate across diverse deployment scenarios. 
    (e) Granular success rate comparison across all tasks and baselines. 
    (f) Cross-embodiment transfer performance evaluated on platforms including wheeled robot, humanoid, and drone. 
    (g) Ablation results illustrating the impact of key component removals across task categories.}
    \label{fig:quantitative_result}
\end{figure}

\subsection{Main Results}
\label{sec:main_results}

\subsubsection{Qualitative Evaluation in Open-World Environments}
\label{sec:qualitative_results}

\textbf{History Backtracking Navigation.} To evaluate the persistent history backtracking capability of \modelname, we conduct both open-loop and closed-loop experiments. 

For open-loop evaluation, we utilize nearly \textit{two-minute} trajectories from the HBN evaluation dataset. 
As illustrated in~\Cref{fig:open_loop}, \modelname accurately and consistently identifies the keyframe index and the direction corresponding to that keyframe near the trajectory origin during the backtracking stage. As progresses, these estimates are dynamically updated upon target re-observation, ensuring precise and stable navigation. 
To the best of our knowledge, this is the first embodied navigation foundation model capable of memory backtracking over nearly two minutes of visual history.

For closed-loop real-world execution shown in Figures~\ref{fig:closed_loop}(a) and (b), \modelname successfully executes a 180$^\circ$ turn to backtrack toward adjacent targets. 
Crucially, even when turning around leaves the target completely out of the current field-of-view, \modelname still achieves successful backtracking by leveraging its long-horizon spatiotemporal reasoning over historical memory.

\textbf{Beyond-the-View Navigation.} Figures~\ref{fig:closed_loop}(c) and (d) demonstrate the target-driven and demand-driven exploration capabilities of \modelname in real-world environments. 
In target-driven tasks, even when navigating into a dead end, \modelname executes a 180$^\circ$ turnaround to exit the confined space and successfully locate the target, effectively avoiding left-right turning deadlocks. 
In demand-driven tasks, \modelname accurately interprets abstract user demands and grounds them to the corresponding target. 

\textbf{Instruction-Following Navigation.} Figures~\ref{fig:closed_loop}(e) to (h) illustrate the instruction following capabilities of \modelname across multi-stage navigation, long straight-line navigation, basic two-stage navigation, and initial-sight target navigation. 
In multi-stage navigation, \modelname accurately grounds the current task stage and unambiguously aligns the turning intention with the language instruction, eliminating directional confusion. 
Upon reaching the final target, \modelname actively makes the decision to stop without relying on heuristic norm-threshold stopping rules, which often trigger premature stopping near obstacles due to reduced action magnitudes.
In long straight-line navigation, \modelname follows forward instructions even when targets remain invisible over extended intervals, facilitated by the history guidance embedded in HRM (see~\Cref{sec:ablation_studies}).
For basic two-stage navigation requiring turns exceeding 90$^\circ$, even if the target remains out-of-view due to an incomplete initial turn, \modelname refrains from random wandering and instead continues the turn under instruction guidance until the large-angle maneuver is fully accomplished.
Finally, in initial-sight target navigation, where the target is unaligned with the initial heading, \modelname continuously adjusts its direction through continuous action sequences and stops precisely in front of the target.

Moreover, \modelname generalizes in a zero-shot manner to both indoor and outdoor scenarios. 
It reliably withstands severe visual distractors, including dynamic human flows crossing the visual field and moving background vehicles, ensuring robust navigation execution.

\subsubsection{Quantitative Comparison with Baselines}
\label{sec:quantitative_results}

\Cref{fig:quantitative_result}(a) presents the comprehensive quantitative comparisons between \modelname and existing SOTA baselines. 
Averaged over four distinct scenarios and eight task configurations, \modelname achieves a dominant overall success rate of $79.4\%$, whereas the second-best baseline AwareVLN attains only $17.8\%$. Other prominent SOTA methods designed for VLN-CE yield drastically lower performance, registering $11.6\%$, $6.3\%$, and $0.3\%$. 
These results strongly indicate that existing SOTA approaches trained purely in simulation suffer from severe overfitting. 
More significantly, fine-tuning these baselines with our dataset fails to yield performance gains comparable to \modelname, demonstrating that high benchmark scores on traditional VLN-CE do not translate to robust open-world deployment capabilities.

\noindent\textbf{Performance across Task Categories.} \Cref{fig:quantitative_result}(b) and (c) break down the success rates by individual tasks and categories. 
For highly challenging HBN, the strongest baseline achieves a mere $8\%$ success rate restricted to basic backtracking, while dropping to $0\%$ on persistent backtracking. 
This reveals that existing SOTA approaches remain fundamentally insensitive to history input even when trained on our dataset, validating that conventional history integration mechanisms are largely ineffective. 
In contrast, \modelname leverages effective history representation and utilization to achieve a $98\%$ success rate on basic backtracking and an unparalleled overall leap from $4\%$ to $78\%$ on HBN. 

In BVN, the strongest baseline exhibits comparable performance between target-driven and demand-driven tasks, suggesting that existing SOTA approaches inherently possess a basic capability to align natural language demands with object categories, albeit constrained by a low performance ceiling. 
\modelname attains an aggregate success rate of $84\%$ in this category, representing a $250\%$ improvement over the strongest baseline. 

For IFN, the strongest baseline achieves approximately $50\%$ success in initial-sight target and basic two-stage navigation. 
However, their performance drops sharply when facing challenging long straight-line and multi-stage tasks, plummeting to $0\%$ in multi-stage navigation. 
Conversely, \modelname achieves over $90\%$ in initial-sight target and basic two-stage navigation, $73\%$ in long straight-line navigation, and $50\%$ in multi-stage navigation, yielding an overall IFN success rate of $78\%$ which is $3.0 \times$ that of the strongest baseline.

\noindent\textbf{Performance across Real-World Scenarios.} \Cref{fig:quantitative_result}(d) illustrates the aggregated success rates across different deployment scenarios. 
Both the strongest baseline and our method maintain relatively stable performance in indoor scenarios and outdoor scenarios with minimal dynamic disturbances. 
However, in densely populated street environments characterized by heavy dynamic disturbances, the performance of the strongest baseline drops by nearly $65\%$. 
Meanwhile, \modelname maintains superior anti-interference capabilities, exhibiting only a minor performance drop.

\noindent\textbf{Failure Case Analysis per Task and Baseline.} \Cref{fig:quantitative_result}(e) provides a granular breakdown of success rates across all task types and baseline models, allowing for an in-depth analysis of failure modes:

\begin{itemize}
    \item \textbf{Initial-Sight Target and Basic Two-Stage Navigation:} Approaches with discrete action outputs, such as AwareVLN and NaVIDA, tend to output straight actions over prolonged distances before executing sharp turns toward targets. While AwareVLN demonstrates superior turning capabilities compared to NaVIDA, this strategy often causes the robot to get blocked by walls after turning in indoor environments, resulting in target occlusion and navigation failure. Although LightNav-0 also utilizes discrete actions, it approximates continuous control for smoother turns. However, control-induced drift during locomotion frequently causes it to deviate and lose the target, such as when traversing stairs. WNM-3D struggles to stop when the target is out of the center of the current field-of-view, highlighting the limited semantic understanding of current world-model-based approaches. Additionally, these three baselines are heavily impacted by dense dynamic disturbances and struggle to execute further turning adjustments when an initial turn is insufficient to align with the target. In contrast, \modelname maintains stable performance, with remaining failures primarily stemming from target occlusion caused by dynamic obstructions or excessive rotation that leads to selected keyframes missing the target.
    
    \item \textbf{Long Straight-Line Navigation:} Approaches with discrete action outputs, such as AwareVLN and NaVIDA, persistently output forward actions during initial stages, which ultimately lead to navigation failure either through uneven terrain and accumulated control errors that cause severe trajectory drift, or via premature stopping in the middle of paths. LightNav-0 frequently exhibits premature turning behaviors that lead to task failure. Although WNM-3D outputs continuous actions and adjusts its heading dynamically via predicted $\text{dy}$, it still fails to maintain a consistent heading direction. While \modelname occasionally exhibits unintended turns during long straight-line navigation, the frequency is significantly lower than that of LightNav-0.
    
    \item \textbf{Multi-Stage Navigation:} Baseline models consistently suffer from various forms of premature stopping and target loss induced by turning inaccuracies. \modelname experiences occasional navigation failures similarly stemming from accumulated turning inaccuracies that cause target loss, particularly in scenarios where targets blend into complex backgrounds due to glass reflections or extreme lighting variations.
    
    \item \textbf{Beyond-the-View Navigation:} LightNav-0 exhibits a high probability for premature turning behaviors, struggling to follow instructions along exploratory paths (e.g., go along the way to find the target), and can directly collide with obstacles in dead ends. AwareVLN and WNM-3D are prone to aggressive exploratory turns that frequently trigger collisions or lead to left-right turning deadlocks. Although AwareVLN generates reasoning text and passes it to the next inference step, its reasoning primarily emphasizes global scene understanding while lacking awareness of previously explored directions. NaVIDA exhibits virtually no active exploration capability. In contrast, \modelname may occasionally drift from the target direction or experience collisions during exploration, though its collision rate remains significantly lower than all comparative baselines. Additionally, it may exhibit transient left-right oscillation behaviors, but typically manages to escape from such loops and resume navigation.
    
    \item \textbf{History Backtracking Navigation:} Among all baselines, only AwareVLN exhibits a slight tendency to turn back, yet it fails to achieve a full $180^\circ$ turn, succeeding only by coincidence when objects happen to be visible with a $120^\circ$ turn. All baseline models completely fail at persistent backtracking. The sole failure mode observed for \modelname on basic tracking involves minor collisions with intermediate targets during U-turn execution. Persistent backtracking remains challenging, where occlusions by obstacles during the return path can cause the robot to lose its return heading despite knowing the approximate target location.
\end{itemize}

\subsubsection{Quantitative Evaluation on Cross-Embodiment Transfer}
\label{sec:diverse_robots}

\Cref{fig:teaser} and~\Cref{fig:quantitative_result}(f) illustrate the qualitative and quantitative evaluation of \modelname across four distinct embodiments under varying camera heights on HBN tasks. 
\modelname achieves robust navigation performance across all evaluated embodiments and camera configurations. 
Due to the inherent motion stability of wheeled robot at the control level, the success rate on this platform is relatively higher. 
Conversely, the unstable pose variations induced by humanoid locomotion and the aerodynamic turbulence during drone flight introduce significant dynamic jitter into captured images, presenting severe OOD visual challenges that slightly degrade performance on these two embodiments.

\subsection{Ablation Studies}
\label{sec:ablation_studies}
\begin{figure}[!ht]
    \centering
    \includegraphics[width=\linewidth]{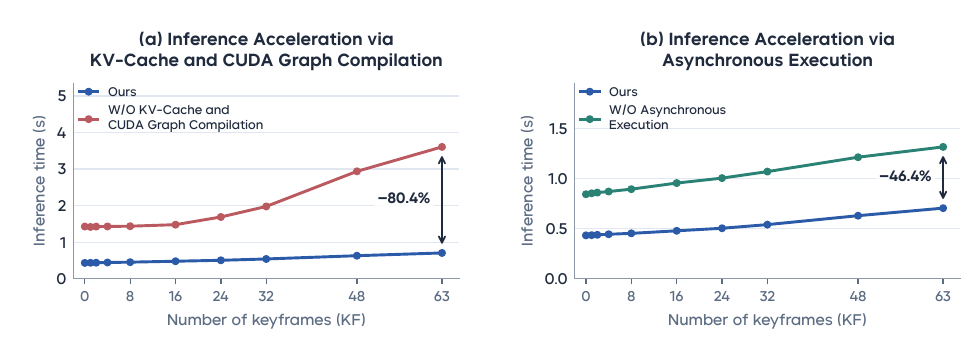}
    \caption{\textbf{Inference efficiency analysis on NVIDIA PRO 6000.} 
    (a) Latency reduction of KV-cache and CUDA graph compilation achieving an $80.4\%$ decrease in execution time at $63$ keyframes. 
    (b) Latency reduction of asynchronous single-GPU execution yielding a $46.4\%$ efficiency gain over synchronous execution across keyframe sequences.}
    \label{fig:inference_efficiency}
\end{figure}

To analyze the internal mechanisms and validate the design choices of \modelname, we conduct comprehensive ablation studies across six variants as illustrated in~\Cref{fig:quantitative_result}(g).

\begin{itemize}
    \item \textbf{w/o Task-Stage Grounding:} 
    This variant removes task-stage grounding VQA and passively determines episode termination based on the norm of action outputs. To preserve intention propagation under this setup, continuous action predictions are discretized and passed in natural language format. This removal causes noticeable performance degradation in basic two-stage navigation due to premature stopping, while dropping performance to zero on multi-stage navigation. This confirms that task-stage grounding VQA provides a critical capability for multi-stage instruction-following. Furthermore, lacking this capability negatively affects other tasks reliant on multi-stage execution, (e.g., persistent backtracking navigation) though its overall impact on exploration and basic backtracking capabilities remains minimal.

    \item \textbf{w/o Exploration Consistency:} 
    This variant removes the exploration consistency VQA and intention propagation. Performance evaluations indicate that this removal primarily impairs target-driven navigation and demand-driven navigation tasks, whereas performance across remaining evaluations stays largely unchanged.

    \item \textbf{w/o Spatiotemporal Target Retrieval:} 
    This variant removes the spatiotemporal target retrieval VQA. Removing this mechanism severely degrades history reasoning capabilities, rendering input HRM completely ineffective without this explicit reasoning trigger. Crucially, this QA mechanism benefits not only history-dependent backtracking navigation, but also mitigates error accumulation in multi-stage navigation. When the model inadvertently miss targets during execution, this spatiotemporal reasoning enables target relocation, thereby increasing overall navigation success rates.

    \item \textbf{w/o HRM:} 
    This variant removes HRM as an extra visual input while retaining all structured VQA tasks. Without HRM, the model preserves partial reasoning capacities, achieving strong performance on basic two-stage navigation alongside reasonable success on multi-stage and basic backtracking navigation. However, when deployed on highly challenging scenarios such as long straight-line and persistent backtracking navigation, relying solely on VQA proves insufficient to satisfy the memory demands. 

    \item \textbf{w/o Video Generation:} 
    This variant removes the inverse dynamics model auxiliary objective and directly attaches a diffusion action head for trajectory generation. Results demonstrate that predictive video generation can improve action generation, proving particularly beneficial for challenging exploration, multi-stage, and history-backtracking navigation.

    \item \textbf{w/o YouTube Data:} 
    This variant restricts model training exclusively to our collected real-world dataset. Co-training with YouTube web-scale video data provides general performance improvements across tasks, though the magnitude of these gains remains constrained by the inherent limited quality of web data.
\end{itemize}

Among the three proposed structured VQA categories, task-stage grounding serves as the foundational bedrock, exploration consistency provides auxiliary guidance, and spatiotemporal target retrieval catalyzes transformational performance gains. Together, these three capabilities form a flexible plug-and-play modular paradigm for visual language navigation.

Comparing HRM against VQA reveals that structured VQA provides the essential foundation for leveraging memory, while HRM provide complementary enhancements for expanding the performance boundary.

\begin{table*}[!ht]
\centering
\caption{\textbf{Comparison on the R2R Val-Unseen split in continuous environments.} S.RGB, Pano., Depth, and Odom. indicate single-view RGB, panoramic or multi-camera RGB, depth, and odometry inputs, respectively. * denotes methods using a waypoint predictor. In monocular setting, \textbf{bold} denotes the best performance.}
\label{tab:r2r_ce}
\resizebox{0.9\textwidth}{!}{%
\begin{tabular}{lcccccccc}
\toprule
\textbf{Method} & \multicolumn{4}{c}{\textbf{Observation}} & \multicolumn{4}{c}{\textbf{R2R Val-Unseen}} \\
\cmidrule(lr){2-5} \cmidrule(lr){6-9}
& S.RGB & Pano. & Depth & Odom. & $\text{NE}\downarrow$ & $\text{OS}\uparrow$ & $\text{SR}\uparrow$ & $\text{SPL}\uparrow$ \\
\midrule
\multicolumn{9}{l}{\textit{Multi-view methods}} \\
$\text{CMA}^*$~\citep{krantz2020beyond} & & $\checkmark$ & $\checkmark$ & $\checkmark$ & 6.20 & 52.0 & 41.0 & 36.0 \\
$\text{HPN+DN}^*$~\citep{krantz2021waypoint} & & $\checkmark$ & $\checkmark$ & $\checkmark$ & 6.31 & 40.0 & 36.0 & 34.0 \\
$\text{Sim2Sim}^*$~\citep{krantz2022sim} & & $\checkmark$ & $\checkmark$ & $\checkmark$ & 6.07 & 52.0 & 43.0 & 36.0 \\
$\text{DreamWalker}^*$~\citep{wang2023dreamwalker} & & $\checkmark$ & $\checkmark$ & $\checkmark$ & 5.53 & 59.0 & 49.0 & 44.0 \\
$\text{HNR}^*$~\citep{wang2024lookahead} & & $\checkmark$ & $\checkmark$ & $\checkmark$ & 4.42 & 67.0 & 61.0 & 51.0 \\
InstructNav~\citep{instructnav2025} & & $\checkmark$ & $\checkmark$ & & 6.89 & $-$ & 31.0 & 24.0 \\
ABot-N0~\citep{chu2026abot} & & $\checkmark$ & & & 3.80 & 70.8 & 66.4 & 63.9 \\
Qwen-RobotNav-4B~\citep{zhang2026qwen} & & $\checkmark$ & & & 3.80 & 77.2 & 69.5 & 63.6 \\
Qwen-RobotNav-8B~\citep{zhang2026qwen} & & $\checkmark$ & & & 3.53 & 78.5 & 72.1 & 66.6 \\
ABot-N1~\citep{gong2026abot} & & $\checkmark$ & & & 3.91 & 71.7 & 68.3 & 66.6 \\
\midrule
\multicolumn{9}{l}{\textit{Monocular methods}} \\
NaVid~\citep{zhang2024navid} & $\checkmark$ & & & & 5.47 & 49.0 & 37.0 & 35.0 \\
Uni-NaVid~\citep{zhang2024uni} & $\checkmark$ & & & & 5.58 & 53.5 & 47.0 & 42.7 \\
NaVILA~\citep{cheng2025navila} & $\checkmark$ & & & & 5.22 & 62.5 & 54.0 & 49.0 \\
SparseVideoNav~\citep{zhang2026sparse} & $\checkmark$ & & & & 4.17 & 56.1 & 48.6 & 47.4 \\
NaVIDA~\citep{navida2026} & $\checkmark$ & & & & 4.32 & 69.5 & 61.4 & 54.7 \\
AwareVLN~\citep{awarevln2026} & $\checkmark$ & & & & 4.02 & 73.5 & 65.4 & 55.1 \\
WNM-3D~\citep{huang2026wnm} & $\checkmark$ & & & & 4.35 & 66.7 & 63.1 & 57.9 \\
LightNav-0~\citep{wang2026lightnav} & $\checkmark$ & & & & 3.91 & 73.7 & 68.5 & 62.8 \\
\rowcolor{cyan!15} \modelname & $\checkmark$ & & & & \textbf{3.88} & \textbf{74.9} & \textbf{70.1} & \textbf{63.2} \\
\bottomrule
\end{tabular}%
}
\end{table*}

\begin{table*}[t]
\centering
\caption{\textbf{Performance on open-vocabulary object navigation.} Comparison on the HM3D-OVON benchmark~\citep{yokoyama2024hm3d} under seen-category, synonyms, and unseen-category settings. In monocular setting, \textbf{bold} denotes the best performance.}
\label{tab:ovon_performance}
\resizebox{0.9\textwidth}{!}{%
\begin{tabular}{lcccccccccc}
\toprule
\textbf{Method} & \multicolumn{4}{c}{\textbf{Observation}} & \multicolumn{2}{c}{\textbf{Seen}} & \multicolumn{2}{c}{\textbf{Synonyms}} & \multicolumn{2}{c}{\textbf{Unseen}} \\
\cmidrule(lr){2-5} \cmidrule(lr){6-7} \cmidrule(lr){8-9} \cmidrule(lr){10-11}
& S.RGB & Pano. & Depth & Odom. & $\text{SR}\uparrow$ & $\text{SPL}\uparrow$ & $\text{SR}\uparrow$ & $\text{SPL}\uparrow$ & $\text{SR}\uparrow$ & $\text{SPL}\uparrow$ \\
\midrule
\multicolumn{11}{l}{\textit{Multi-view methods}} \\
Qwen-RobotNav-4B~\citep{zhang2026qwen} & & $\checkmark$ & & & 57.7 & 24.4 & 60.1 & 25.1 & 53.1 & 20.9 \\
Qwen-RobotNav-8B~\citep{zhang2026qwen} & & $\checkmark$ & & & 56.1 & 28.5 & 57.8 & 28.8 & 51.2 & 24.0 \\
ABot-N0~\citep{chu2026abot} & & $\checkmark$ & & & 55.3 & 32.1 & 55.4 & 33.2 & 54.0 & 30.5 \\
\midrule
\multicolumn{11}{l}{\textit{Monocular methods}} \\
VLFM~\citep{vlfm2024} & $\checkmark$ & & $\checkmark$ & $\checkmark$ & 35.2 & 18.6 & 32.4 & 17.3 & 35.2 & 19.6 \\
DAgRL+OD~\citep{yokoyama2024hm3d} & $\checkmark$ & & $\checkmark$ & $\checkmark$ & 38.5 & 21.1 & 39.0 & 21.4 & 37.1 & 19.8 \\
MTU3D~\citep{zhu2025move} & $\checkmark$ & & $\checkmark$ & $\checkmark$ & 55.0 & 23.6 & 45.0 & 14.7 & 40.8 & 12.1 \\
Uni-NaVid~\citep{zhang2024uni} & $\checkmark$ & & & & 41.3 & 21.1 & 43.9 & 21.8 & 39.5 & 19.8 \\
LightNav-0~\citep{wang2026lightnav} & $\checkmark$ & & & & 55.3 & 31.2 & 54.6 & 29.6 & 47.0 & 24.2 \\
\rowcolor{cyan!15} \modelname & $\checkmark$ & & & & \textbf{57.4} & \textbf{31.3} & \textbf{59.7} & \textbf{31.5} & \textbf{52.2} & \textbf{26.9} \\
\bottomrule
\end{tabular}%
}
\end{table*}

\begin{table*}[t]
\centering

\begin{minipage}[t]{0.47\textwidth}
\centering
\captionof{table}{\textbf{Comparison results on IntentionNav.} \textbf{Bold} denotes the best performance and \underline{underline} denotes the second best performance.}
\label{tab:intentionnav}
\resizebox{0.9\textwidth}{!}{%
\begin{tabular}{lcccc}
\toprule
\textbf{Model} & $\text{NE}\downarrow$ & $\text{OS}\uparrow$ & $\text{SR}\uparrow$ & $\text{SPL}\uparrow$ \\
\midrule
NaVIDA~\citep{navida2026} & 2.53 & 42.5 & 33.5 & 33.3 \\
WNM-3D~\citep{huang2026wnm} & 2.99 & 43.0 & 30.0 & 29.9 \\
AwareVLN~\citep{awarevln2026} & 3.15 & \textbf{60.0} & 33.5 & 29.8 \\
LightNav-0~\citep{wang2026lightnav} & \textbf{2.28} & 38.5 & \underline{35.5} & \underline{35.1} \\
\rowcolor{cyan!15} \modelname & \underline{2.35} & \underline{50.5} & \textbf{42.5} & \textbf{42.1} \\
\bottomrule
\end{tabular}%
}
\end{minipage}%
\hfill 
\begin{minipage}[t]{0.5\textwidth}
\centering
\captionof{table}{\textbf{Comparison results on IndoorUAV-VLN~\citep{liu2026indooruav} test unseen split.} \modelname and all baselines are fine-tuned on the dataset. \textbf{Bold} denotes the best performance and \underline{underline} denotes the second best performance. $*$ denotes the checkpoint open-sourced by IndoorUAV.}
\label{tab:indoor_uav}
\resizebox{0.85\textwidth}{!}{%
\begin{tabular}{lcccc}
\toprule
\textbf{IndoorUAV-VLN} & \multicolumn{4}{c}{\textbf{TEST UNSEEN}} \\
\cmidrule(lr){2-5}
& $\text{NE}\downarrow$ & $\text{OS}\uparrow$ & $\text{SR}\uparrow$ & $\text{nDTW}\uparrow$ \\
\midrule
LLaMA-UAV~\citep{wang2025towards} & 8.49 & 4.26 & 3.19 & 4.70 \\
OpenFly~\citep{gao2026openfly} & \underline{7.06} & 10.64 & 4.26 & \underline{11.88} \\
$\pi_0^*$~\citep{black2025pi0} & 10.53 & \textbf{43.62} & \underline{9.57} & 5.85 \\

\rowcolor{cyan!15} \modelname & \textbf{5.94} & \underline{37.23} & \textbf{18.09} & \textbf{12.14} \\
\bottomrule
\end{tabular}%
}
\end{minipage}

\end{table*}
\subsection{Inference Efficiency Analysis}
\label{sec:inference_efficiency}

We evaluate the inference efficiency of \modelname across varying keyframe lengths as quantitative metrics presented in~\Cref{fig:inference_efficiency}. 
Compared to the version without KV-cache and CUDA graph compilation, our optimization pipeline achieves an $80.4\%$ reduction in inference latency at $63$ keyframes. Furthermore, as the keyframe count increases, our latency exhibits a remarkably slow growth trend, demonstrating robust context scaling capabilities. 
Simultaneously, introducing the asynchronous single-GPU execution strategy yields a speedup of nearly $50\%$ over synchronous execution. 
These optimizations boost the deployment responsiveness, enabling our policy to compute $12$ continuous actions in $0.43\text{s}$ at the beginning, which corresponds to an effective control frequency of approximately $28\text{Hz}$, and maintain $0.71\text{s}$ latency at $63$ keyframes, representing approximately $17\text{Hz}$.

\subsection{Simulation Evaluation Results}
\label{sec:simu_results}

We further evaluate the performance of \modelname using established benchmarks. 
Note that HBN currently lacks a simulation benchmark for comprehensive evaluation.
We report navigation error (NE), oracle success rate (OS), success rate (SR), and success weighted by path length (SPL) as evaluation metrics.

\begin{itemize}
    \item \textbf{R2R-CE~\citep{anderson2018vision} (for IFN):} We train \modelname from scratch using R2R with DAgger augmentation. As shown in~\Cref{tab:r2r_ce}, \modelname achieves SOTA performance among existing monocular navigation algorithms.
    
    \item \textbf{OVON~\citep{yokoyama2024hm3d} (for target-driven navigation of BVN):} We train \modelname from scratch using HM3D-OVON data and augment the training with DAgger. As shown in~\Cref{tab:ovon_performance}, \modelname achieves SOTA performance among existing monocular navigation algorithms.
    
    \item \textbf{IntentionNav~\citep{qian2026intentionnav} (for demand-driven navigation of BVN):} IntentionNav serves as a demand-driven navigation benchmark designed to evaluate whether the policy can locate the target based on user demands. Since the data source in IntentionNav is entirely unseen by all evaluated methods, we partition its 500 trajectory samples into a training set of 300 trajectories and a testing set of 200 trajectories. These 300 training trajectories are utilized to fine-tune \modelname and all baselines without DAgger augmentation. Detailed records of trajectory split and baseline implementations are documented in~\Cref{sec:app_impl_simu}. As shown in~\Cref{tab:intentionnav}, \modelname achieves SOTA performance on IntentionNav.

    \item \textbf{IndoorUAV~\citep{liu2026indooruav} (for IFN):} IndoorUAV is a multi-stage benchmark that closely aligns with indoor drone flight altitudes. We remove all trajectories with up and down actions from both training and evaluation set, and then feed the training trajectories into \modelname and baselines for fine-tuning, followed by evaluation on the evaluation set. Since IndoorUAV does not provide ground-truth shortest trajectories, we report Normalized Dynamic Time Warping (nDTW) instead of SPL. \Cref{tab:indoor_uav} shows that \modelname achieves SOTA performance on the more challenging multi-stage IFN task.
\end{itemize}

\section{Related Work}
\label{sec:related_work}

\noindent
\textbf{History Representation in Embodied Navigation Foundation Models.}
Existing end-to-end foundation models predominantly process visual history through three paradigms: feeding dense uncompressed streams or sliding windows into the backbone~\citep{zhang2024navid, huang2026wnm}, applying heuristic token merging and multi-scale memory pooling~\citep{zhang2024uni, wang2026lightnav, torne2026memmultiscaleembodiedmemory}, or uniformly subsampling frames into a fixed context window~\citep{cheng2025navila, awarevln2026}. 
However, these strategies either flood attention with redundant tokens or drop critical spatial information~\citep{zhang2026sparse}, rendering HBN unachievable. 
Although recent frameworks like MemoryVLN~\citep{memoryvln} and MapNav~\citep{mapnav} have introduced explicit memory banks and semantic maps as additional inputs, the policy tends to largely ignore them due to the lack of explicit supervision, making them still insufficient for HBN.

\vspace{4pt}
\noindent
\textbf{VQA Supervision in Embodied Navigation.}
To bridge high-level semantic reasoning and low-level motion control, recent methods increasingly incorporate VQA and CoT supervision during training. 
Within embodied navigation, \citet{lin2025navcot} and \citet{liu2025navr1} supervise policies with step-by-step navigational CoT traces, while \citet{wang2026vlingnav} and \citet{awarevln2026} dynamically trigger interleaved reasoning tokens to switch between the reasoning of scene and action generation. 
Similarly, \citet{wang2026lightnav} and \citet{navida2026} augment policy training with latent point reasoning and action-grounded visual dynamics. 
However, these auxiliary formulations predominantly target general scene understanding, future prediction, or coarse subgoal identification~\citep{rgpt2024, zhang2026qwen, zhang2026supernav}, leaving history reasoning largely underexplored and lacking the mechanism to analyze and resolve closed-loop physical execution failures~\citep{huang2026wnm, zhang2026sparse}. 

\vspace{4pt}
\noindent
\textbf{Learning from Real-World Navigation Videos and Datasets.}
Large-scale learning-based navigation policies have historically relied on synthetic trajectories scaled within photorealistic simulators such as Habitat~\citep{2019habitat, vlnce2020, wang2023scalevln}. 
Despite claims of zero-shot sim-to-real transfer~\citep{zhang2024uni, cheng2025navila, huang2026wnm}, simulator-trained policies suffer from visual artifacts and physical domain gaps that degrade real deployment. 
While real-world robot datasets such as ViNT~\citep{shah2023vint} and NoMaD~\citep{sridhar2024nomad} aggregate multi-embodiment physical trajectories, they focus on point-goal or image-goal navigation and lack fine-grained language annotations. 
To harvest open-world visual priors with minimal collection overhead, \citet{hirose2024lelan} and \citet{urbannav} annotate in-the-wild YouTube videos, while \citet{navdreamer2026} extract metric pseudo-trajectories via 3D vision foundation models~\citep{wang2026pi, 2025moge2}. 
However, these pipelines yield only single-object labels or unverified motion primitives without the stage-aligned instructions needed for embodied navigation. 
Recently, \citet{zhang2026sparse} demonstrated strong physical deployment via real-world data collection, yet their framework relies on labor-intensive labeling and lacks VQA supervision. Consequently, it is crucial to develop a lightweight data labeling pipeline that minimizes manual effort.

\section{Conclusion}

In this work, we present \modelname, an embodied navigation foundation model designed to pursue open-world and persistent navigation capabilities with nearly two minutes of history backtracking.
By integrating keyframe-level sparsification with an explicit HRM, our method effectively preserves long-horizon spatiotemporal context without introducing architectural modifications. 
Complementing this, the proposed structured VQA supervision protocol serves as an essential map activation mechanism, successfully mitigating pervasive real-world execution failures and enforcing history-aware reasoning across instruction-following, beyond-the-view, and history backtracking navigation tasks. 
Furthermore, our semi-automated dataset curation pipeline significantly reduces human overhead and enables the scalable acquisition of diverse multi-modal interaction data. 
Extensive real-world and simulation evaluations demonstrate that \modelname achieves superior performance and robust zero-shot generalization across diverse physical scenarios and robotic embodiments. Coupled with detailed failure case analyses, our work provides comprehensive insights into closed-loop execution and lays a solid foundation for the future development of navigation embodied systems.


\clearpage

\bibliographystyle{plainnat}
\bibliography{main}

\clearpage

\beginappendix

\section{Implementation and Reproducibility Details for Baselines}
We here present the implementation details for fine-tuning baseline models on real-world data during open-world evaluations (\Cref{sec:app_impl_real}), as well as the training and testing data split of IntentionNav (\Cref{sec:app_impl_simu}).

\begin{figure}[ht]
    \centering
    \begin{subfigure}[t]{0.48\linewidth}
        \centering
        \includegraphics[width=\linewidth]{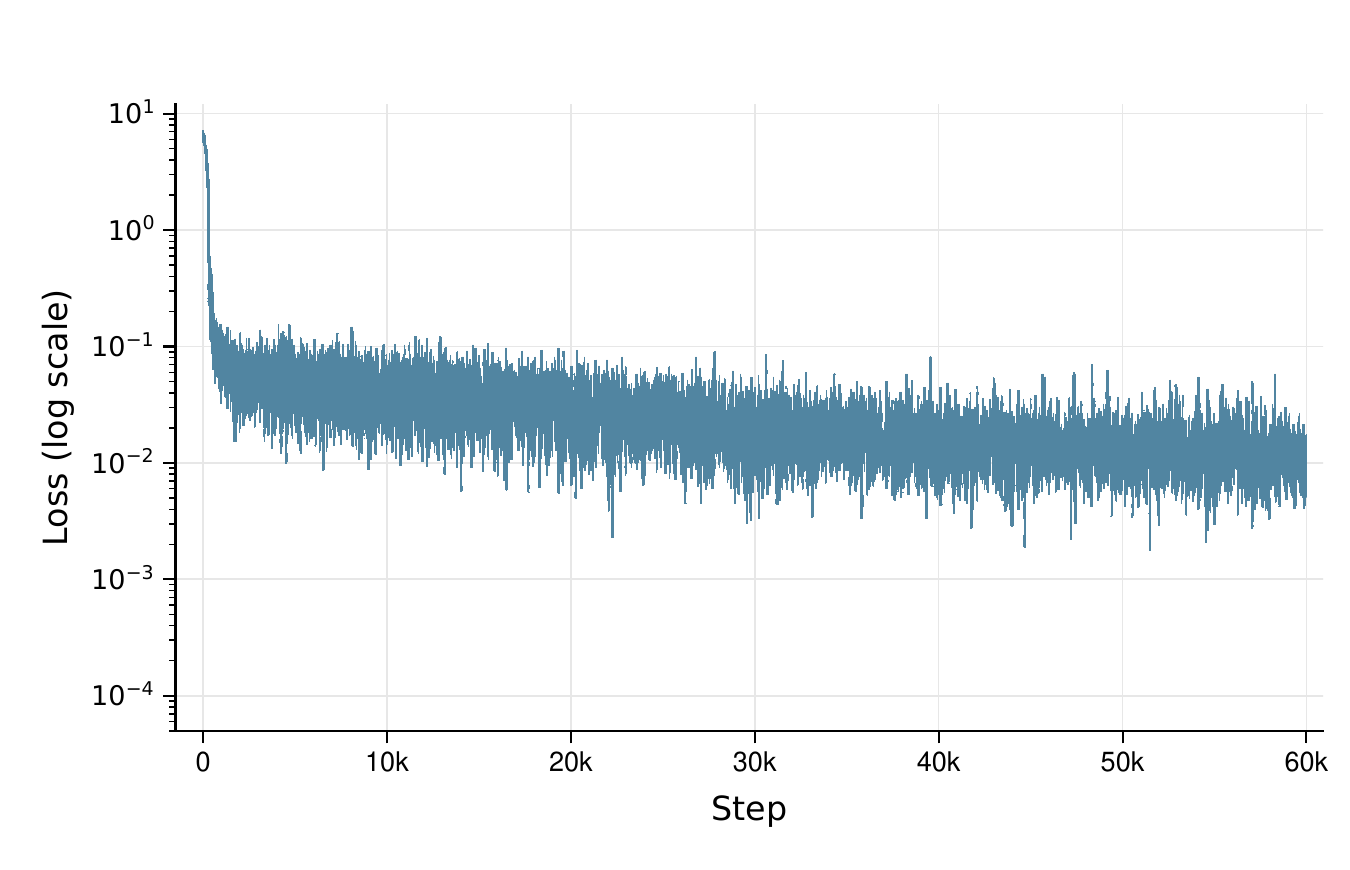}
        \caption{AwareVLN}
        \label{fig:loss_aware_real}
    \end{subfigure}
    \hfill
    \begin{subfigure}[t]{0.48\linewidth}
        \centering
        \includegraphics[width=\linewidth]{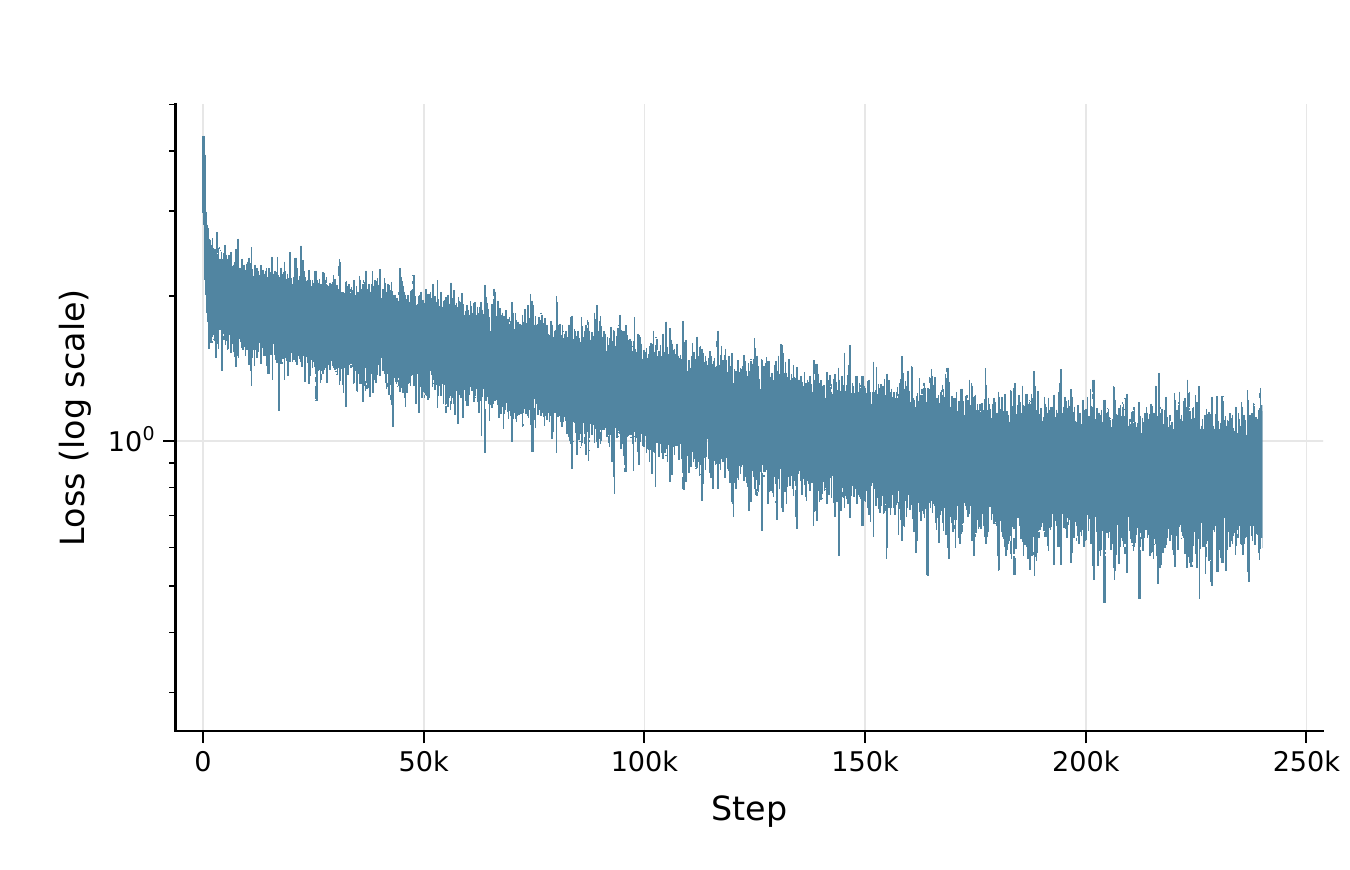}
        \caption{LightNav-0}
        \label{fig:loss_light_real}
    \end{subfigure}

    \medskip

    \begin{subfigure}[t]{0.48\linewidth}
        \centering
        \includegraphics[width=\linewidth]{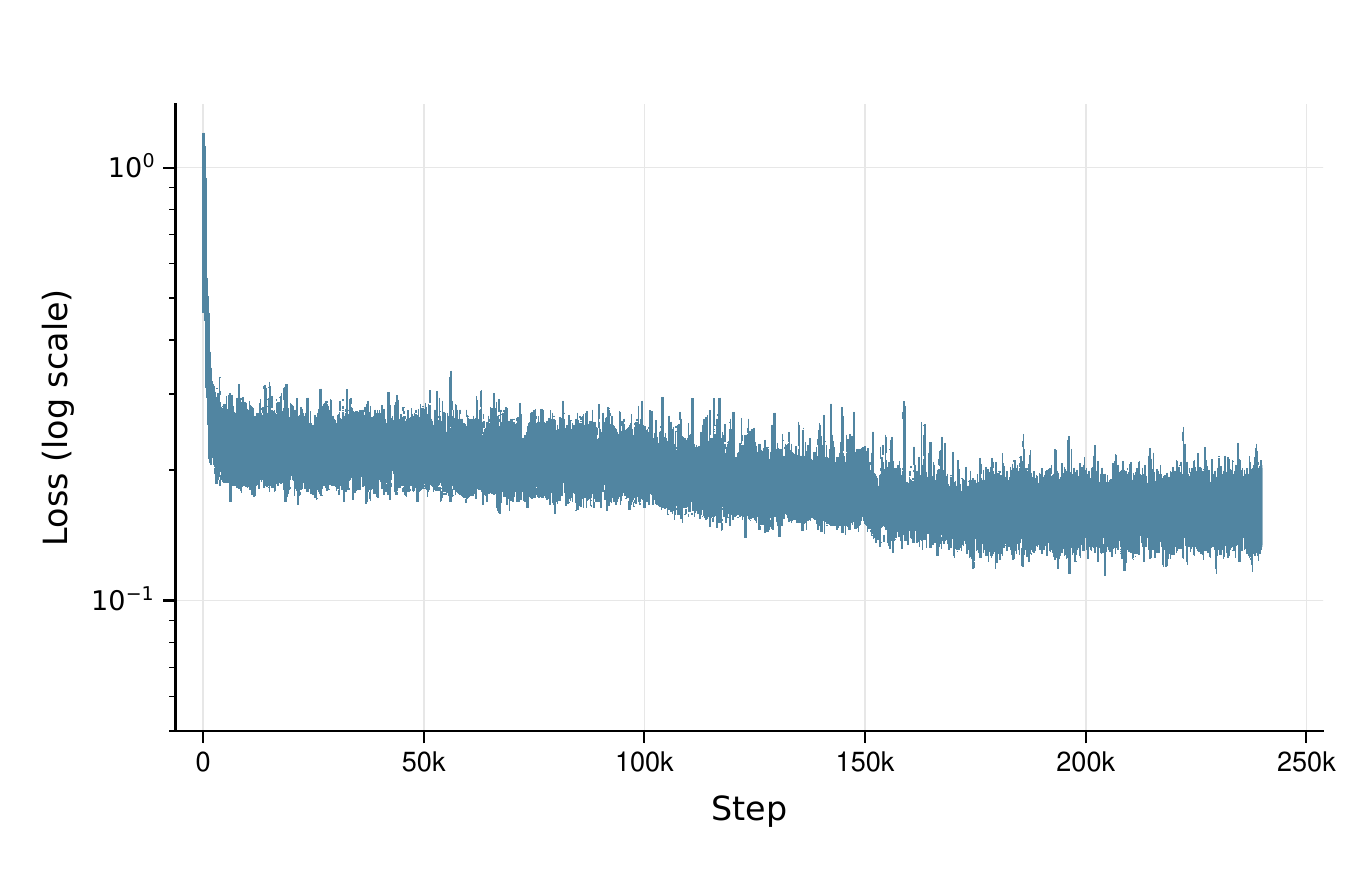}
        \caption{NaVIDA}
        \label{fig:loss_navida_real}
    \end{subfigure}
    \hfill
    \begin{subfigure}[t]{0.48\linewidth}
        \centering
        \includegraphics[width=\linewidth]{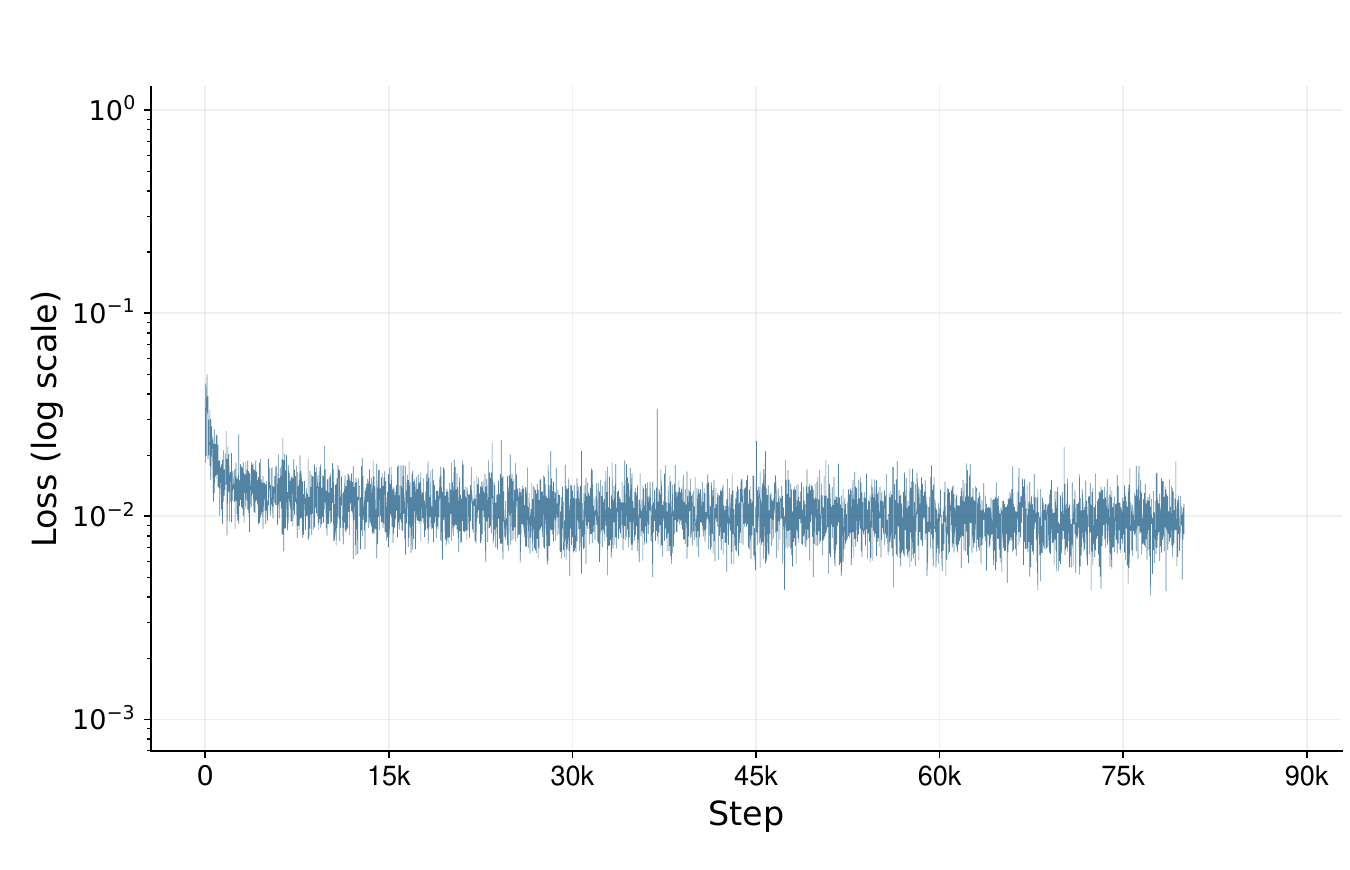}
        \caption{WNM-3D}
        \label{fig:loss_wnm_real}
    \end{subfigure}

    \caption{\textbf{Training loss curves for baseline fine-tuning on real-world data.}
    NaVIDA, AwareVLN, and LightNav-0 report autoregressive cross-entropy loss, while WNM-3D reports action flow-matching loss.}
    \label{fig:baseline_real_loss}
\end{figure}

\subsection{Implementation and Reproducibility Details for Baselines in Real-World}
\label{sec:app_impl_real}

We initialize all four baselines from their officially released checkpoints and adapt them to our real-world data following their original training frameworks.
To reflect the peak performance of each baseline, we report the higher success rate between its original open-sourced weight and its fine-tuned counterpart in our evaluation results.
We detail the baseline implementations as following:

\textbf{AwareVLN~\citep{awarevln2026}.}
Based on normalized forward displacement and rotation magnitude, we map continuous real-world motions to four action categories: moving forward by 25\,cm, turning left by $15^\circ$, turning right by $15^\circ$, and stopping.
The model takes historical and current RGB observations together with the task instruction as input.
Since reasoning outputs require additional text annotations, we apply cross-entropy loss only to the action output, without additional reasoning supervision.
We reduce the global batch size from 768 in the original configuration to 256 and empirically adjust the learning rate from $2\times10^{-5}$ to $1\times10^{-5}$ and find it yields better performance.
To ensure a fair comparison, we employ the same number of training samples as in the training setup of \modelname.
As shown in~\Cref{fig:baseline_real_loss}(a), the training of AwareVLN approaches near-convergence around $60\text{,}000$ steps.
We find that despite the absence of direct supervision on reasoning outputs during fine-tuning, AwareVLN successfully preserves its pre-existing reasoning abilities.

\textbf{LightNav-0~\citep{wang2026lightnav}.}
Since depth information is unavailable in real-world videos, we approximate the affordance point by using camera parameters to project the action position at the 4th future step onto the image plane. 
As for the goal point, it can be directly obtained by taking the midpoint of the target's grounding bounding box.
We retain the original autoregressive cross-entropy objective and jointly supervise the pointing-of-affordance token, pointing-of-goal-object token, three RVQ action tokens, and end-of-sequence token.
We reduce the global batch size from 320 to 64 and empirically find that setting the learning rate to $1\times10^{-5}$ yields better performance.
To maintain the same number of training samples of \modelname, we train LightNav-0 for $240\text{,}000$ steps.
As shown in~\Cref{fig:baseline_real_loss}(b), the training curve exhibits a clear convergence trend.

\textbf{NaVIDA~\citep{navida2026}.}
We discretize continuous actions following the identical mapping protocol of AwareVLN. 
We then use HPAC to process the discrete action sequences, merging adjacent actions of the same class with a probability of $0.7$, which strictly follows the official open-source configuration.
Under this setup, we further aggregate the continuous action in the same type to make sure each action block contains at most three action segments (e.g., turn left, turn right, forward).
Fine-tuning is conducted using the original global batch size of 64 and the default learning rate for 240k steps to match the total training samples of \modelname. 
As shown in~\Cref{fig:baseline_real_loss}(c), the training curve exhibits a clear convergence trend.

\textbf{WNM-3D~\citep{huang2026wnm}.}
To ensure a fair comparison, we maintain a global batch size of 192 and the original learning rate, training for $80\text{,}000$ steps to match the training setup of \modelname.
Since the model natively supports continuous actions, we directly align three-dimensional actions with RGB videos without discretization.
As shown in~\Cref{fig:baseline_real_loss}(d), the training curve exhibits a clear convergence trend.

\subsection{Training and Testing Data Split of IntentionNav}
\label{sec:app_impl_simu}

Due to the limited scale of IntentionNav data, we leverage Qwen3-VL~\citep{bai2025qwen3} to generate reasoning data for all IntentionNav scenes to fine-tune AwareVLN on this benchmark. Meanwhile, for LightNav-0, since depth maps are available in the simulator, Qwen3-VL is utilized to generate 2D candidate points based on multi-view RGB and instructions. 
Then, depth and camera poses are retrieved to verify the geometric consistency of these candidate points for subsequent projection onto the 2D image plane.

The official IntentionNav~\citep{qian2026intentionnav} contains 500 intention-driven trajectories.
We use the same split across all four baselines, randomly sampling 300 training trajectories and 200 testing trajectories within each scene such that every environment contains both training and test tasks.
The complete list of the 200 evaluation trajectory indices are provided in
\Cref{tab:ddn_eval_ids}.

\begin{table}[H]
\centering
\caption{\textbf{Complete testing trajectory indices of IntentionNav.} Note that the trajectory indices are non-consecutive.}
\label{tab:ddn_eval_ids}
\small
\setlength{\tabcolsep}{5pt} 
\begin{tabular}{*{15}{c}}
\toprule
003 & 004 & 005 & 008 & 009 & 017 & 018 & 020 & 021 & 023 & 025 & 029 & 035 & 037 & 039 \\
044 & 045 & 047 & 048 & 051 & 060 & 061 & 065 & 069 & 071 & 072 & 074 & 075 & 076 & 079 \\
083 & 086 & 089 & 091 & 092 & 093 & 094 & 095 & 097 & 098 & 099 & 100 & 101 & 102 & 104 \\
107 & 109 & 112 & 113 & 117 & 118 & 120 & 121 & 123 & 134 & 137 & 140 & 144 & 146 & 147 \\
150 & 151 & 153 & 155 & 157 & 163 & 174 & 176 & 179 & 180 & 186 & 187 & 194 & 196 & 198 \\
199 & 209 & 210 & 212 & 213 & 214 & 215 & 218 & 219 & 225 & 230 & 233 & 235 & 237 & 241 \\
243 & 249 & 252 & 257 & 258 & 259 & 260 & 261 & 262 & 265 & 266 & 267 & 270 & 274 & 277 \\
278 & 279 & 281 & 282 & 283 & 285 & 293 & 295 & 296 & 300 & 304 & 306 & 310 & 316 & 318 \\
319 & 320 & 321 & 324 & 326 & 328 & 334 & 335 & 336 & 338 & 341 & 344 & 345 & 352 & 358 \\
360 & 362 & 364 & 367 & 372 & 373 & 374 & 378 & 379 & 380 & 384 & 385 & 388 & 390 & 392 \\
393 & 396 & 398 & 403 & 404 & 405 & 407 & 410 & 412 & 415 & 417 & 422 & 423 & 424 & 425 \\
426 & 427 & 431 & 446 & 448 & 449 & 451 & 453 & 454 & 455 & 459 & 460 & 461 & 464 & 467 \\
468 & 475 & 477 & 479 & 481 & 487 & 494 & 495 & 503 & 506 & 509 & 510 & 511 & 512 & 513 \\
516 & 518 & 519 & 521 & 524 & & & & & & & & & & \\
\bottomrule
\end{tabular}
\end{table}

\section{Open-World Benchmark Configuration}
\label{sec:app_openworld_config}

We here present the comprehensive suite of open-world evaluation task specifications across four representative deployment environments. 
In total, 32 distinct navigation tasks are defined, encompassing instruction-following navigation (IFN), Beyond-the-View Navigation (BVN), and History-Backtracking Navigation (HBN).

\begin{figure}[!ht]
    \centering
    \includegraphics[width=\linewidth]{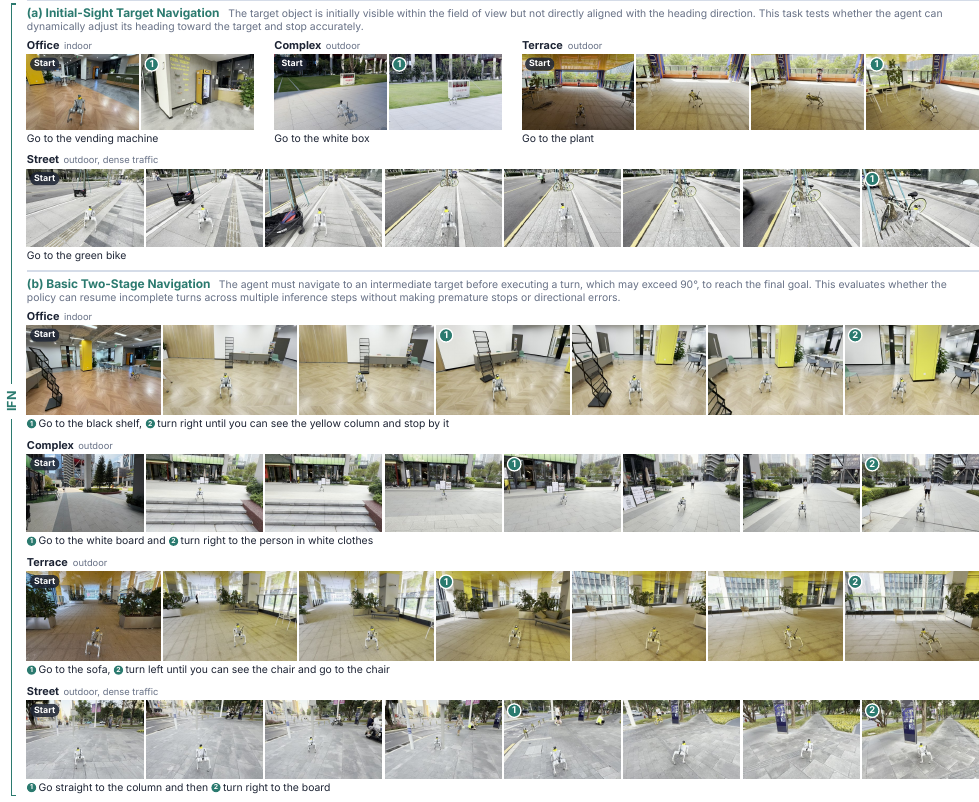}
    \caption{\textbf{Task specifications for Initial-Sight Target Navigation and Basic Two-Stage Navigation of IFN} across office, complex, terrace, and street environments.}
    \label{fig:appendix_tasks_p1}
\end{figure}
\begin{figure}[!ht]
    \centering
    \includegraphics[width=\linewidth]{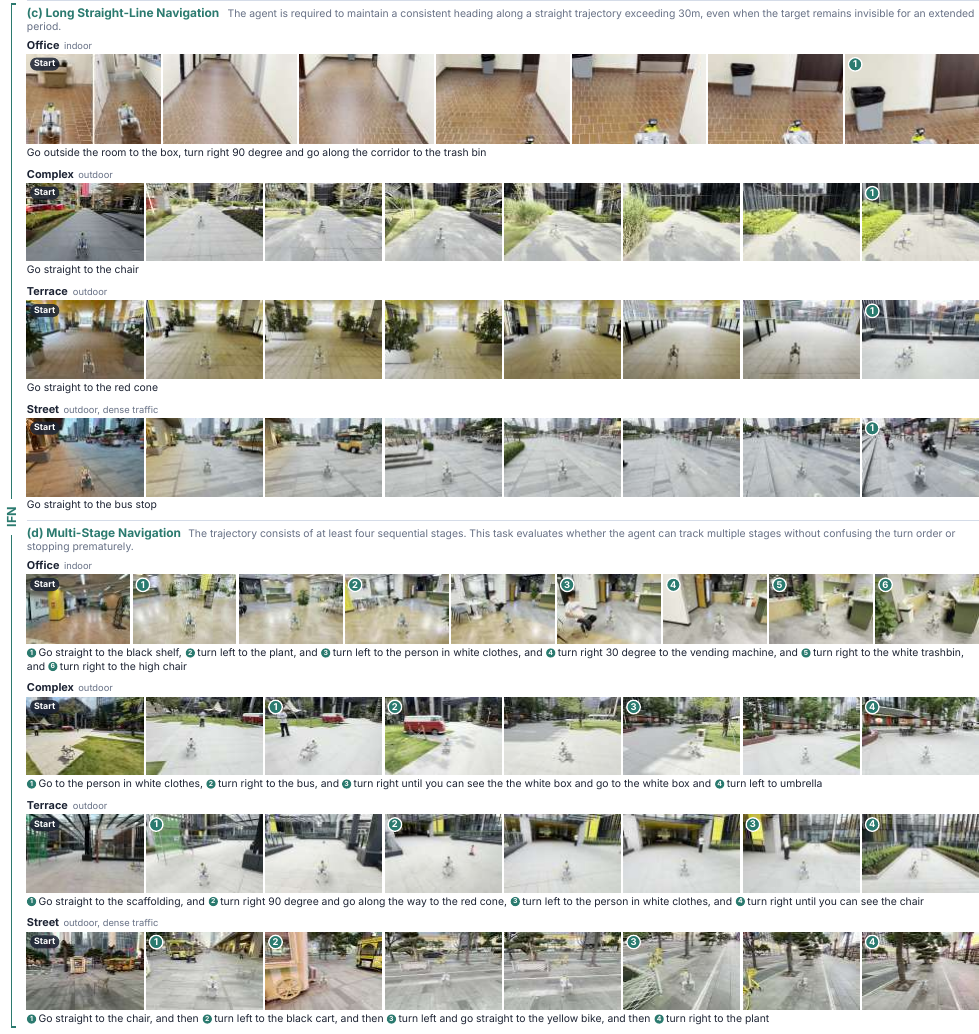}
    \caption{\textbf{Task specifications for Long Straight-Line Navigation and Multi-Stage Navigation of IFN} across office, complex, terrace, and street environments.}
    \label{fig:appendix_tasks_p2}
\end{figure}
\begin{figure}[!ht]
    \centering
    \includegraphics[width=\linewidth]{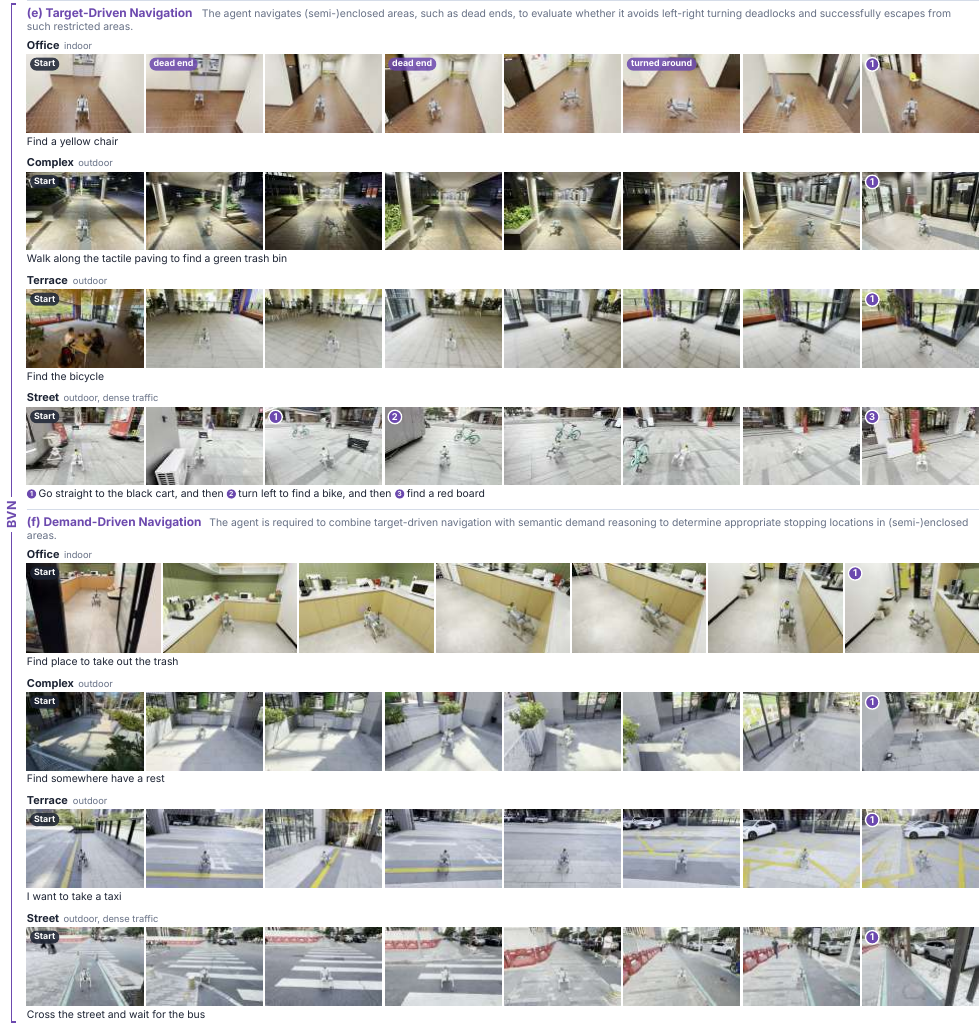}
    \caption{\textbf{Task specifications for Target-Driven Navigation and Demand-Driven Navigation of BVN} across office, complex, terrace, and street environments.}
    \label{fig:appendix_tasks_p3}
\end{figure}
\begin{figure}[!ht]
    \centering
    \includegraphics[width=\linewidth]{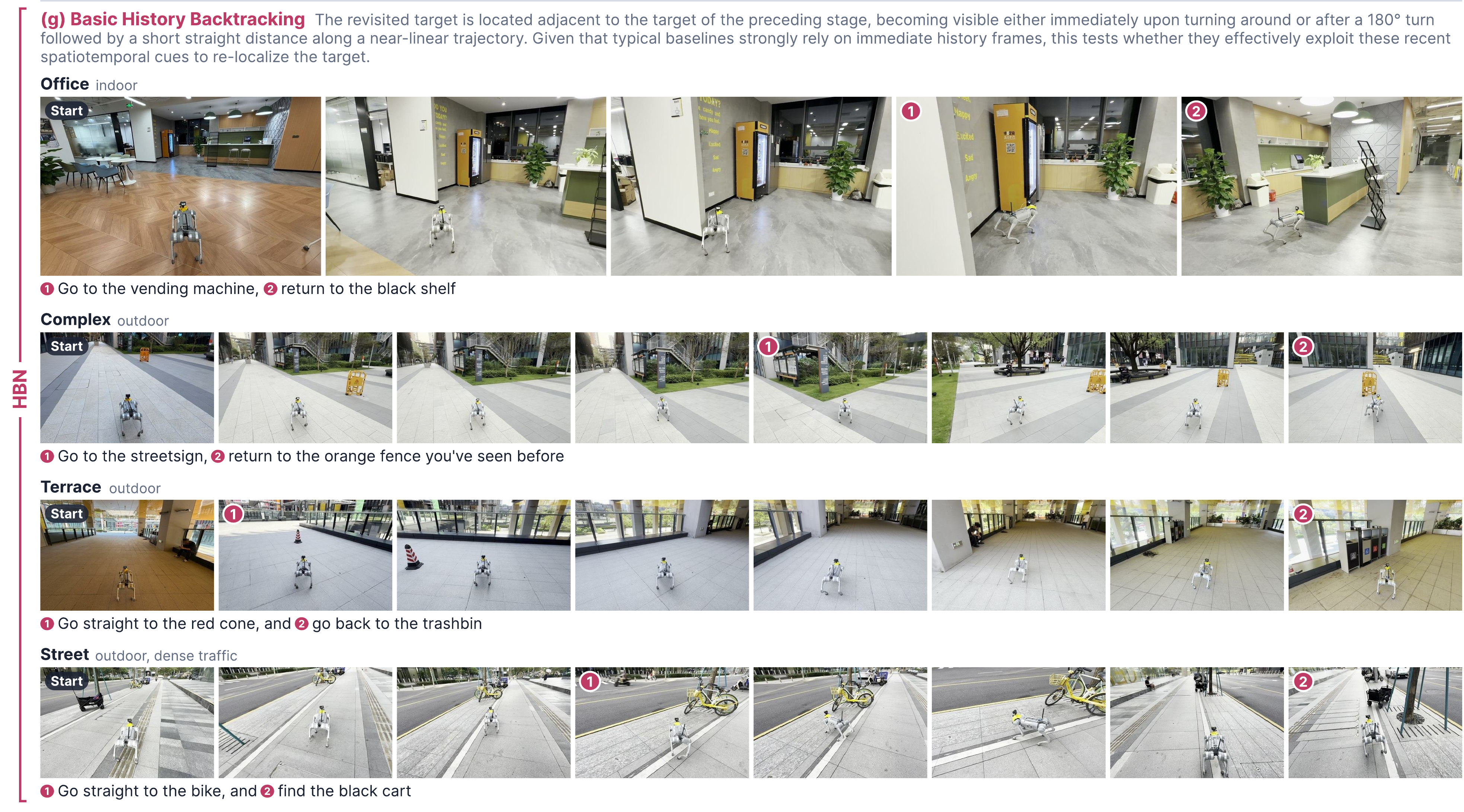}
    \caption{\textbf{Task specifications for Basic History Backtracking of HBN} across office, complex, terrace, and street environments.}
    \label{fig:appendix_tasks_p4}
\end{figure}
\begin{figure}[!ht]
    \centering
    \includegraphics[width=\linewidth]{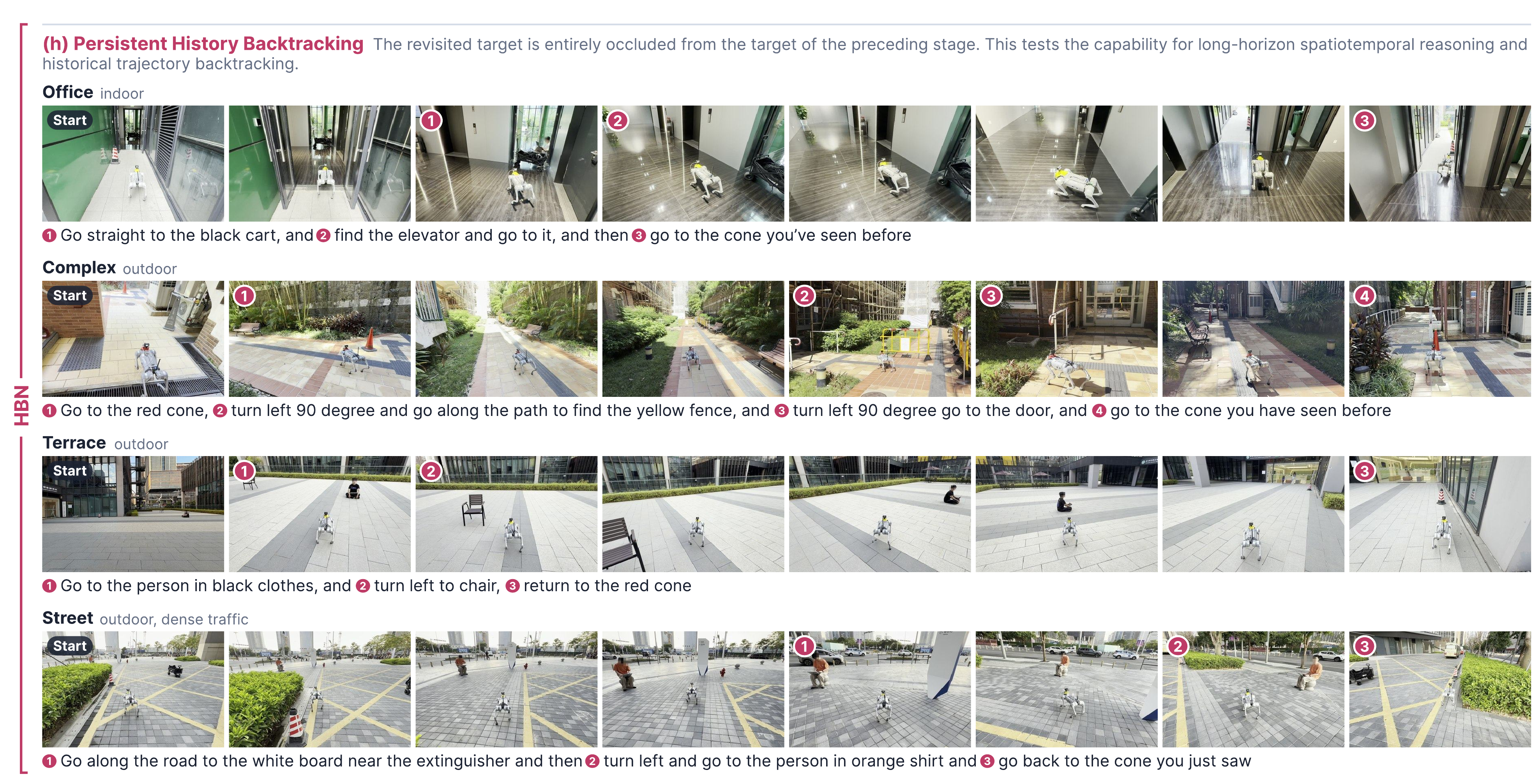}
    \caption{\textbf{Task specifications for Persistent History Backtracking of HBN} across office, complex, terrace, and street environments.}
    \label{fig:appendix_tasks_p5}
\end{figure}

\end{document}